\documentclass[10pt,twocolumn,letterpaper]{article}

\usepackage[pagenumbers]{cvpr} 

\usepackage{blindtext}
\usepackage{duckuments}
\usepackage{multirow}

\usepackage{pifont}
\definecolor{Gray}{gray}{0.9} 

\newcommand\blfootnote[1]{
    \begingroup
    \renewcommand\thefootnote{}\footnote{#1}
    \addtocounter{footnote}{-1}
    \endgroup
}

\definecolor{cvprblue}{rgb}{0.21,0.49,0.74}
\usepackage[pagebackref,breaklinks,colorlinks,allcolors=cvprblue]{hyperref}
\newcommand{\paragrapht}[1]{\noindent\textbf{#1}}

\newcommand{\ours }{C3G-$\mathcal{G}$}
\newcommand{\oursF }{C3G-$\mathcal{F}$}
\newcommand{\oursname }{\textbf{C3G}}

\def\paperID{583} 
\def\confName{CVPR}
\def\confYear{2026}

\title{C3G: Learning Compact 3D Representations with 2K Gaussians}

\author{
    Honggyu An\textsuperscript{\rm 1}$^{*}$ \quad
    Jaewoo Jung\textsuperscript{\rm 1}$^*$ \quad
    Mungyeom Kim\textsuperscript{\rm 1} \quad
    Chaehyun Kim\textsuperscript{\rm 1}  \\
    Minkyeong Jeon\textsuperscript{\rm 1} \quad
    Jisang Han\textsuperscript{\rm 1} \quad
    Kazumi Fukuda\textsuperscript{\rm 3} \quad
    Takuya Narihira\textsuperscript{\rm 3}$^{\dagger}$ \\
    Hyunah Ko\textsuperscript{\rm 1} \quad
    Junsu Kim\textsuperscript{\rm 1} \quad
    Sunghwan Hong\textsuperscript{\rm 2}$^{\dagger}$ \quad
    Yuki Mitsufuji\textsuperscript{\rm 3,4}$^{\dagger}$ \quad
    Seungryong Kim\textsuperscript{\rm 1}$^{\dagger}$ \\[10pt]
    \textsuperscript{\rm 1}KAIST AI \qquad \textsuperscript{\rm 2}ETH AI Center, ETH Z\"urich \qquad \textsuperscript{\rm 3}SONY AI \qquad \textsuperscript{\rm 4}Sony Group Corporation\\
{\tt \href{https://cvlab-kaist.github.io/C3G}{\textcolor{purple}{https://cvlab-kaist.github.io/C3G}}}
}

\begin{document}

\twocolumn[{%
\renewcommand\twocolumn[1][]{#1}%
\maketitle

\begin{center}
    \centering
    \captionsetup{type=figure}
    \includegraphics[width=\linewidth]{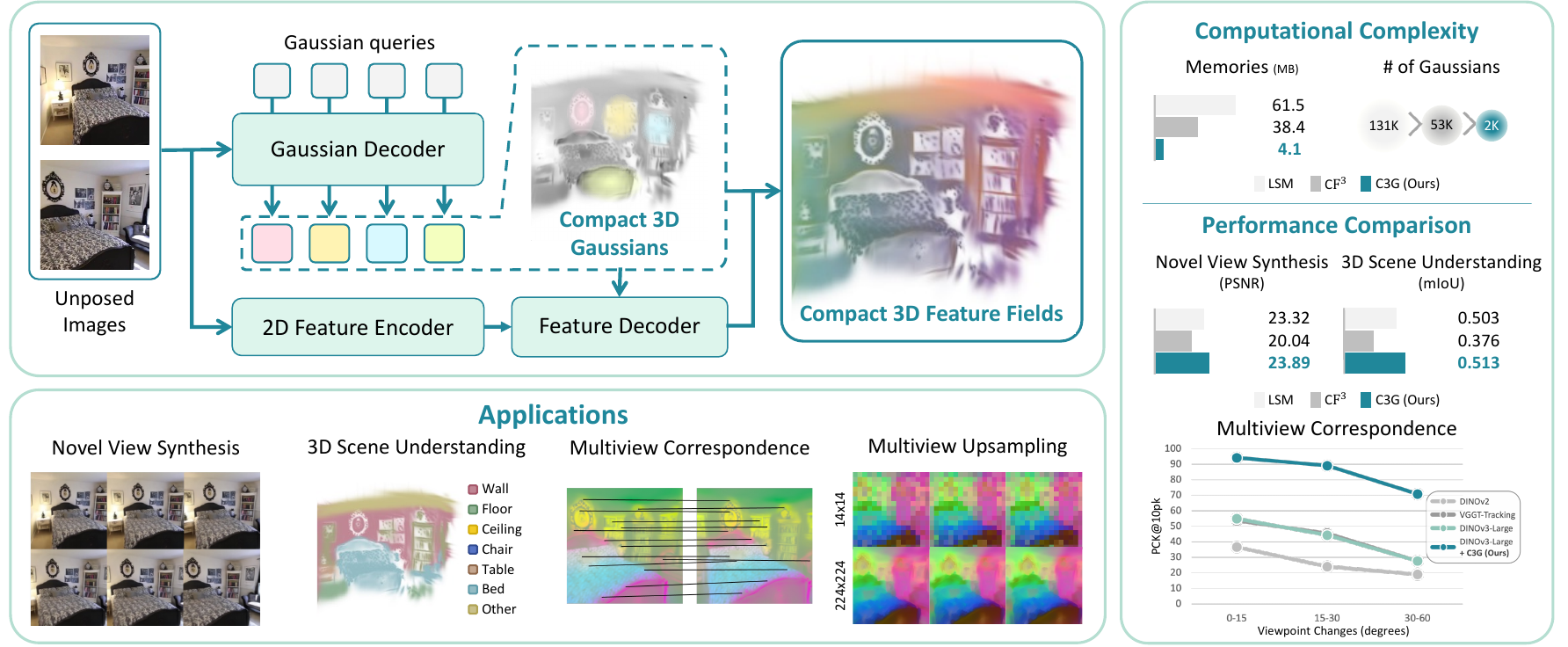}
    \vspace{-20pt}
    \captionof{figure}{\textbf{Teaser.} Our method learns compact 3D Gaussians from unposed multi-view images through a query-based Gaussian decoding pipeline. Compact representations enable efficient 2D-to-3D feature lifting for downstream applications, including 3D understanding, correspondence, and upsampling. Compared to prior works~(LSM~\cite{fan2024large} and CF$^3$~\cite{lee2025cf3}), Our \textbf{C3G} results in the fewest Gaussians—about \textbf{2K}, which is roughly \textbf{65$\times$} fewer than LSM~\cite{fan2024large}—while achieving superior memory efficiency and novel view synthesis quality.}
    \label{fig:teaser}
\end{center}
}]
\blfootnote{$^*$Equal contributions.}
\blfootnote{$^\dagger$Co-advising.}

\begin{abstract}
Reconstructing and understanding 3D scenes from unposed sparse views in a feed-forward manner remains as a challenging task in 3D computer vision. Recent approaches use per-pixel 3D Gaussian Splatting for reconstruction, followed by a 2D-to-3D feature lifting stage for scene understanding. However, they generate excessive redundant Gaussians, causing high memory overhead and sub-optimal multi-view feature aggregation, leading to degraded novel view synthesis and scene understanding performance. We propose \textbf{C3G}, a novel feed-forward framework that estimates compact 3D Gaussians only at essential spatial locations, minimizing redundancy while enabling effective feature lifting. We introduce learnable tokens that aggregate multi-view features through self-attention to guide Gaussian generation, ensuring each Gaussian integrates relevant visual features across views. We then exploit the learned attention patterns for Gaussian decoding to efficiently lift features. Extensive experiments on pose-free novel view synthesis, 3D open-vocabulary segmentation, and view-invariant feature aggregation demonstrate our approach's effectiveness. Results show that a compact yet geometrically meaningful representation is sufficient for high-quality scene reconstruction and understanding, achieving superior memory efficiency and feature fidelity compared to existing methods.
\end{abstract}
\vspace{-10pt}    
\section{Introduction}
\label{sec:intro}
Obtaining 3D representations from unposed sparse multi-view images in a feed-forward manner remains a fundamental challenge in computer vision and graphics, with broad implications for robotics~\cite{shen2023distilled,han2025d}, scene understanding~\cite{fedele2025superdec}, and novel view synthesis~\cite{ye2024no}. Recently, feed-forward 3D Gaussian splatting frameworks have gained considerable attention, demonstrating impressive performance in reconstruction and understanding~\cite{ye2024no,huang2025no,hong2024pf3plat,fan2024large,sheng2025spatialsplat}.

\begin{figure*}[t]
    \centering
    \includegraphics[width=\linewidth]{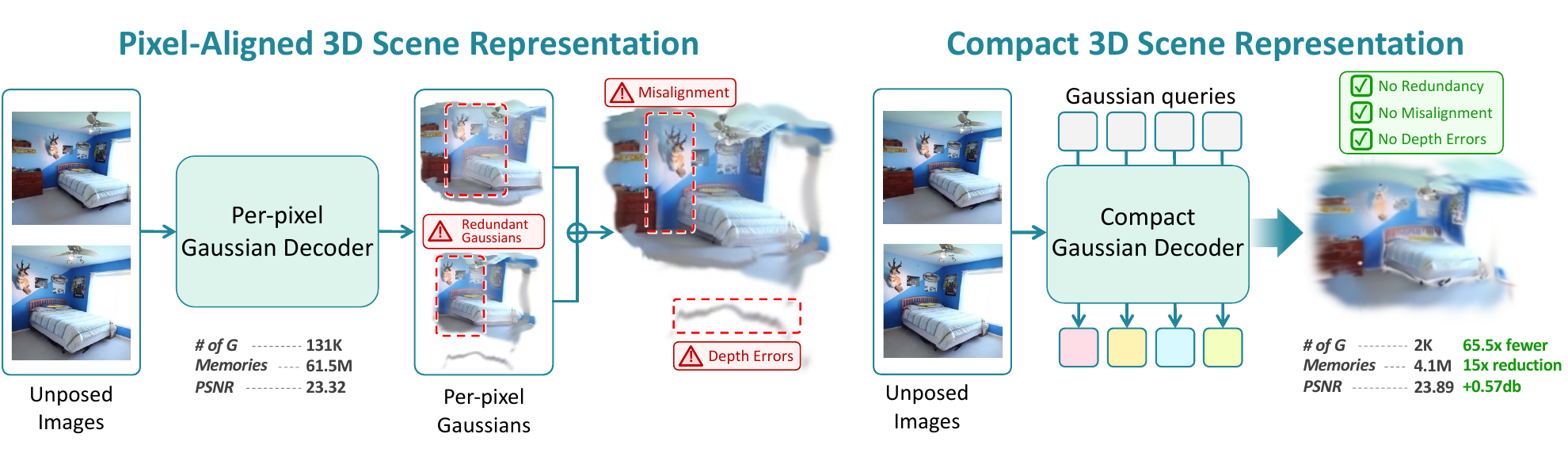}
    \vspace{-20pt}
    \caption{\textbf{Comparison of per-pixel and compact scene representations.} \textbf{(Left)}: Existing per-pixel estimators~\cite{ye2024no,huang2025no} predict one or multiple Gaussians per pixel, resulting in redundant Gaussians with misalignments across views. \textbf{(Right)}: Our method uses learnable Gaussian queries to discover and decode only compact 3D Gaussians at essential locations, achieving a compact representation with only 2K Gaussians and 4.1M memory while avoiding redundancy and achieving superior segmentation and novel view synthesis performance.}
    \label{fig:concept}
    \vspace{-10pt}
\end{figure*}

However, these approaches predominantly rely on \textit{dense, per-pixel} Gaussian predictions, which inherently leads to two critical issues: (1) it generates excessive redundant primitives that are often misaligned in 3D space~(Fig.~\ref{fig:concept}), and (2) it incurs substantial computational overhead when incorporating semantic features~\cite{sheng2025spatialsplat,fan2024large} through 2D-to-3D feature lifting. Consequently, prior works compress rich semantic information~\cite{simeoni2025dinov3,oquab2023dinov2,kim2025seg4diff} into lower-dimensional embeddings, resulting in information loss and sub-optimal scene understanding. This raises a fundamental question: \textbf{\textit{do we need such pixel-aligned Gaussians to reconstruct and understand 3D scenes?}}

As humans, we do not maintain pixel-perfect mental reconstructions of every surface to understand our surroundings. Instead, we form compact, semantically meaningful abstractions of identifying key objects, their rough spatial relationships, and overall scene structure~\cite{burgess2006spatial,shepard1971mental, lee2025perspective}. Drawing direct inspiration from human visual cognition, we propose a novel framework \oursname\ for learning \textbf{\textit{compact 3D Gaussians}} from unposed image observations in a feed-forward fashion. 

Similar to prior approaches~\cite{ye2024no,huang2025no}, we first extract image features from visual encoders with rich geometric priors (e.g., VGGT~\cite{wang2025vggt}). However, instead of learning to estimate per-pixel Gaussians directly from the extracted feature maps, we introduce a compact set of learnable query tokens that discover and decode essential 3D Gaussians. Specifically, we adopt a transformer architecture~\cite{vaswani2017attention} where learnable query tokens and the image features are processed through multiple self-attention blocks. We decode the refined learnable query tokens as 3D Gaussians, where the query tokens learn to aggregate essential information across multiple views to faithfully represent the scene.

Crucially, our framework requires no explicit supervision from ground-truth depths or scene decompositions. Despite training solely on photometric reconstruction, each token naturally learns to represent different regions, with each token attending to coherent spatial regions across views. This emergent behavior arises from the inherent structure of the task: to efficiently reconstruct novel views with a limited number of Gaussians, the model must learn to allocate Gaussians to meaningful regions. 

We show that after training, our model can estimate a compact set of 3D Gaussians, which enables efficient novel view synthesis while maintaining performance. In addition, the compact set of Gaussians enables 2D-to-3D feature lifting without compression, significantly improving 3D scene understanding tasks where the rich representation of the semantic features is critical.

Our framework also provides a novel solution to a key challenge in 2D-to-3D feature lifting: handling multi-view feature inconsistencies. While other methods~\cite{zeid2025dino, lee2025cf3, cheng2024occam, marrie2025ludvig} require additional aggregation methods to handle inconsistent features across viewpoints, we observe that our model's emergent property of tokens attending to spatially coherent regions across views can be directly leveraged for feature aggregation. Specifically, we propose a view-invariant feature decoder~(\textbf{\oursF}) that reuses the attention maps from our learned Gaussian decoder~(\textbf{\ours}) while training only the value projections. This feature decoder can then take features from any visual encoder as input and decode multi-view aggregated features. By attaching these aggregated features to our estimated 3D Gaussians, we enable efficient novel view rendering with view-invariant features.

We validate the effectiveness of our approach through extensive experiments. For novel view synthesis, despite using \textbf{65$\times$} fewer Gaussians (only \textbf{2K}) than per-pixel methods~\cite{ye2024no,huang2025no}, we achieve competitive visual quality while enabling substantially faster rendering. More importantly, for 3D semantic understanding tasks, our compact Gaussians combined with the multi-view aggregated semantic features significantly outperform previous feed-forward approaches that attempt to lift view-inconsistent features through dense per-pixel Gaussians~\cite{fan2024large}. We further verify that our feature renderings can effectively replace previous feature upsamplers, while also outperforming previous view-invariant feature aggregation methods in multi-view correspondence evaluation.
\section{Related work}
\label{sec:relwork}

\paragrapht{Learning compact 3D scene representations.} 
Decomposing scenes into geometric primitives~(e.g., meshes, polygons, superquadric primitives) has been extensively studied~\cite{tulsiani2017learning, paschalidou2019superquadrics, paschalidou2020learning}, but these methods typically require 3D data~(e.g., point clouds) as input. SuperDec~\cite{fedele2025superdec} recently proposed feed-forward decomposition into superquadric primitives, but relies on pre-computed 3D point clouds and requires iterative application of an external 3D instance segmentation model to identify instances. The differentiable blocks world (DBW)~\cite{monnier2023differentiable} learns superquadric parameters directly from multi-view images through photometric optimization, but can only model up to 10 primitives and object-centric scenes. In contrast, we aim to learn an efficient solution that can estimate compact scene representations in a feed-forward manner, only given unposed multi-view images captured in real-world scenarios without any additional ground-truth labels. 

\paragrapht{Feed-forward 3D Gaussian splatting.} 
Recently, 3D Gaussian Splatting (3DGS)~\cite{kerbl20233d} has gained significant attention for 3D reconstruction and novel view synthesis. However, it requires dense multi-view captures with known poses and time-consuming per-scene optimization. To address these limitations, recent approaches~\cite{charatan2024pixelsplat, chen2024mvsplat, du2023learning, johari2022geonerf, wang2021ibrnet, xu2024murf, yu2021pixelnerf, zhang2025transplat} have explored generalizable feed-forward networks~\cite{ye2024no, hong2024unifying, huang2025no, hong2024pf3plat, jiang2025anysplat, smart2024splatt3r} that synthesize novel views from sparse inputs by learning priors from large-scale datasets or leveraging foundation models such as pointmap regression models~\cite{wang2024dust3r, leroy2024grounding, wang2025vggt}. Despite these advances, existing feed-forward models~\cite{ye2024no, huang2025no} typically predict one or multiple Gaussians per-pixel, resulting in millions of Gaussians for multiple views or high-resolution images. This strategy produces excessive redundant Gaussians, degrading performance and causing artifacts as input views increase~\cite{wang2024freesplat, wang2025volsplat}. When incorporating semantic features through 2D-to-3D lifting, the excessive Gaussians create significant computational overhead~\cite{lee2025cf3, burgess2006spatial}. Previous works~\cite{zhou2024feature, fan2024large, burgess2006spatial, qin2024langsplat} address this by compressing semantic features into lower dimensions using specialized autoencoders, but this causes information loss and sub-optimal scene understanding~\cite{li2025langsplatv2}.

\paragrapht{Towards compact 3D Gaussian splatting.}
To address the redundancy problem in per-pixel Gaussian estimation, recent works~\cite{wang2024freesplat, huang2025longsplat, wang2025zpressor, zhang2024gaussian, ziwen2025long,jiang2025anysplat} have attempted to mitigate this issue by reducing the number of Gaussians post-hoc. FreeSplat~\cite{wang2024freesplat} and LongSplat~\cite{huang2025longsplat} iteratively add pixel-wise Gaussians only where existing projections are insufficient. ZPressor~\cite{wang2025zpressor} and Long-LRM~\cite{ziwen2025long} employ token merging to reduce redundant Gaussians with similar features before decoding them to per-pixel Gaussians. Anysplat~\cite{jiang2025anysplat} voxelizes the Gaussians in 3D space. However, these approaches do not fundamentally address the input-view bias inherent in per-pixel processing. EvolSplat~\cite{miao2025evolsplat} and VolSplat~\cite{wang2025volsplat} employ global voxel representations but remain domain-constrained or limited by fixed resolutions. We instead propose using learnable tokens to generate compact global Gaussians guided by input features, producing only essential Gaussians for scene representation.

\begin{figure*}[t]
    \centering
    \includegraphics[width=\linewidth]{figure/assets/figure3_attn.pdf}
    \vspace{-15pt}
    \caption{\textbf{Architecture and emergent attention behaviors of our 3D Gaussian decoder~(\ours).} (a) Our framework first extracts multi-view features using VGGT, then processes them with learnable query tokens through transformer blocks in our Gaussian decoder~(\ours). The refined queries are subsequently decoded into compact 3D Gaussians via a Gaussian head, trained with the novel view synthesis loss~$\mathcal{L}_\text{novel}$.
(b) Visualization of learned attention patterns between a target Gaussian (\textcolor[HTML]{FF0000}{red} dots) and the features in Gaussian-to-Image~(G2I) attention. Without any explicit supervision, each query token learns to attend to spatially coherent regions across multiple views, naturally discovering their corresponding regions.}
    \label{fig:framework}
    \vspace{-10pt}
\end{figure*}

\section{Methodology}
\label{sec:method}
We introduce \ours, a compact 3D Gaussian splatting decoder built upon a transformer~\cite{vaswani2017attention} that takes multi-view images of a scene as input and produces a compact set of 3D Gaussians that best represent the scene. We start by introducing the problem~(\textbf{\S~\ref{subsec:problem_def}}), followed by our architectural details~(\textbf{\S~\ref{subsec:architecture}}), and the training setup~(\textbf{\S~\ref{subsec:training}}). We further analyze the emergent properties in our learned \mbox{\ours~(\textbf{\S~\ref{subsec:analysis}})}, and show how this property can be leveraged to effectively lift any 2D features into 3D in a view-invariant manner~(\textbf{\S~\ref{subsec:feature_lifting}}).

\subsection{Problem definition and notation}
\label{subsec:problem_def}
Given a set of $V$ multi-view images $\{I_v\}_{v=1}^V$ capturing the same scene where $I_v\in\mathbb{R}^{H \times W \times 3}$, the model outputs a set of $N$ 3D Gaussians $\{\mathbf{G}_i\}_{i=1}^N$ with $\mathbf{G}_i=\{\mu_i,\sigma_i,\Sigma_i,c_i\}_{i=1}^N$, where $\mu_i \in\mathbb{R}^3$ denotes 3D Gaussian center, $\sigma_i \in [0,1)$ represents opacity, $\Sigma_i \in \mathbb{R}^{3 \times 3}$ denotes the covariance matrix, and $c_i \in \mathbb{R}^{3(L+1)}$ indicates spherical harmonics coefficients with $L$ levels that encode color attributes. Although other formats of 3D scene representations (e.g., point clouds~\cite{wang2025vggt,wang2024dust3r}, polygons~\cite{nan2017polyfit}) could be considered, we adopt 3D Gaussians~\cite{kerbl20233d} as our default representation due to their efficient rendering speeds and simplicity to further incorporate features as an additional attribute that enables multiple downstream tasks. 

\subsection{Architecture}
\label{subsec:architecture}
Drawing inspiration from how humans naturally form abstract scene representations through selective attention, we design a remarkably simple architecture that learns to decode compact 3D Gaussians from multi-view observations. 
\paragrapht{Multi-view feature encoding.}
To decode the 3D Gaussians, given the $V$ input images $\{I_v\}_{v=1}^V$, we first extract visual features using a pre-trained visual encoder $\mathcal{E}(\cdot)$, yielding feature maps $\mathbf{F}_v \in \mathbb{R}^{h \times w \times d}$ for each view, where $h$ and $w$ are the height and width of the feature map and $d$ is the feature dimension. To effectively encode multi-view images, we follow prior works and adopt VGGT~\cite{wang2025vggt} as our default visual encoder, which has learned rich geometric priors from large-scale geometry learning.

\paragrapht{Query-based scene decoding.} 
At the core of our architecture is the compact set of $N$ learnable query tokens $\mathbf{Q} \in \mathbb{R}^{N \times d}$, where each token is tasked with discovering and representing a specific region of the 3D scene. These tokens serve as abstraction units that learn to aggregate relevant information from the extracted multi-view features $\mathbf{F}_v$ to form coherent 3D Gaussians. Unlike per-pixel Gaussian estimation methods~\cite{ye2024no, charatan2024pixelsplat} that rigidly map each pixel to Gaussians, our query tokens can flexibly attend to any region across all input views, learning to allocate representational capacity where it is most needed.

\paragrapht{Cross-view attention aggregation.}
The key to our approach lies in how query tokens interact with multi-view features. We concatenate the learnable query tokens with the image features to form a unified sequence: \mbox{$\mathbf{X} = [\mathbf{Q}; \mathbf{F}] \in \mathbb{R}^{N + (V \times h \times w) \times d}$}. This combined representation is processed through $L$ transformer layers with full self-attention, enabling bidirectional information flow. Through these attention layers, each query token learns to: (1) aggregate relevant visual information from specific regions across all input views, (2) exchange information with other query tokens to avoid redundancy and ensure comprehensive coverage, and (3) progressively refine its understanding of which 3D region it should represent.

\paragrapht{Gaussian parameter decoding.}
After the transformer blocks, we extract the refined query tokens \mbox{${\bar{\mathbf{Q}}_i} \in \mathbb{R}^{N \times d}$} and decode each token ${\bar{\mathbf{Q}}_i}$ into a single Gaussian $\mathbf{G}_i$, estimating Gaussian attributes through lightweight MLP heads.

\subsection{Training}
\label{subsec:training}
Unlike previous methods that learn scene decompositions from ground-truth labels~\cite{fedele2025superdec,paschalidou2019superquadrics}, adopting 3D Gaussians as our representation enables our framework to be learned solely through the objective of novel view synthesis.

\paragrapht{Learning compact scene representations from novel view synthesis.}
Given the predicted 3D Gaussians $\{\mathbf{G}_i\}_{i=1}^N$ from our query tokens, we train the model by rendering these Gaussians at novel target viewpoints and minimizing the photometric difference with ground-truth images. Following feed-forward novel view synthesis frameworks~\cite{ye2024no}, we project the 3D Gaussians to a target view $I_t \notin \{I_v\}_{v=1}^V$ with known camera pose $\pi_t$ during training. Each pixel ${p}$ of the target view image is rendered via alpha blending of Gaussian color attributes according to their depth order~\cite{kerbl20233d}:
\begin{equation}
\hat{I}_t({p}) = \sum_{i=1}^N {c}_i\sigma_i\mathbf{G}^\text{2D}_i({p}) \prod_{j=1}^{i-1}(1-\sigma_j \mathbf{G}^\text{2D}_j({p})),
\end{equation}
where ${c}_i$ is the view-dependent color attribute of each Gaussian obtained by decoding spherical harmonics coefficients, and $\mathbf{G}^\text{2D}_i$ is the 3D Gaussian projected onto 2D screen space. Our training objective combines the mean squared error between rendered and ground-truth images $\mathcal{L}_{\text{MSE}}$ and the perceptual loss $\mathcal{L}_{\text{LPIPS}}$ as: 
\begin{equation}
\mathcal{L}_\text{novel} = \lambda_{\text{MSE}} \mathcal{L}_{\text{MSE}}(\hat{I}_t,I_t) + \lambda_{\text{LPIPS}} \mathcal{L}_{\text{LPIPS}}(\hat{I}_t,I_t),
\end{equation}
following prior works~\cite{ye2024no}, where $\lambda_{\text{MSE}}$ and $\lambda_{\text{LPIPS}}$ are hyperparameters.

\paragrapht{Low-pass filtering for robust training.} 
One of the key challenges in learning feed-forward 3D Gaussian splatting is correctly locating Gaussian positions~($\mu_i$). Without accurate positions, prior works show that the Gaussians often fail to be positioned inside the view frustum of the target image viewpoint, leading to sparse gradients and mode collapse~\cite{charatan2024pixelsplat, hong2024pf3plat}. Aligned with these analyses, we also observe that naively training the network with photometric loss leads to unstable training. To address this, we adopt the progressive low-pass filter from RAIN-GS~\cite{jung2024relaxing}. For rendering Gaussian $\mathbf{G}_i$, the projected 2D Gaussian $\mathbf{G}^\text{2D}_i$ is defined as follows:
\begin{equation}
   \mathbf{G}^\text{2D}_{i}(p) = e^{-\frac{1}{2}(p-\mu_{i}^\text{2D})^T(\Sigma_{i}^\text{2D}+s\mathbf{I})^{-1}(p-\mu_{i}^\text{2D})},
\end{equation}
where $p$ is the 2D pixel location, $\mathbf{I}$ is the identity matrix, $s$ controls the Gaussian size and $\mu_i^\text{2D}$ and $\Sigma_i^\text{2D}$ denote 2D projected positions and covariance of $\mathbf{G}_i$. While 3DGS~\cite{kerbl20233d} uses $s=0.3$ to ensure 1-pixel coverage, RAIN-GS shows that progressively annealing from $s=300$ to $s=0.3$ stabilizes per-scene optimization by allowing Gaussians to learn from enlarged regions initially. We adopt this strategy in our feed-forward training pipeline, gradually annealing $s$ during training. This ensures robust gradients during early training when position predictions are suboptimal, while enabling fine-grained detail to be modeled as the network learns accurate positions. Our ablations~(\textbf{\S~\ref{subsec:ablation}}) further verify that this strategy is crucial for stable training.

\subsection{Analysis}
\label{subsec:analysis}
Although we do not provide any supervision for how the $N$ query tokens should partition the scene, we observe that the model eventually learns to effectively estimate a set of $N$ Gaussians that best reconstructs the scene purely from the photometric reconstruction objective. 

\paragrapht{Emergent properties within learned attentions.}
To understand how each query token learns to aggregate information from multi-view features, we examine the attention weights between learnable tokens $\mathbf{Q}$ and the multi-view image features $\mathbf{F}_v$ inside the \textit{self-attention blocks} of \mbox{\ours}. As visualized in Fig.~\ref{fig:framework}-(a), we examine the attention weights where the attention query is from the $N$ learnable tokens $Q_G$ and the attention key is from the multi-view image features $K_I$. 

As illustrated in Fig.~\ref{fig:framework}-(b), when we select a specific Gaussian and visualize its corresponding query token's attention map across input views, we observe sharp, focused attention patterns on spatially coherent regions across multiple views, effectively discovering multi-view correspondences without any explicit supervision~\cite{an2025cross}. For instance, the target Gaussian highlighted in red attends strongly to the corresponding object regions across different viewpoints. We believe that this emergent behavior arises from an implicit optimization pressure: to accurately reconstruct novel views with a limited number of $N$ Gaussians, the model has learned to position 3D Gaussians to geometrically coherent regions.

\subsection{Any-feature 3D lifting}
\begin{figure}[t]
    \centering
    \includegraphics[width=\linewidth]{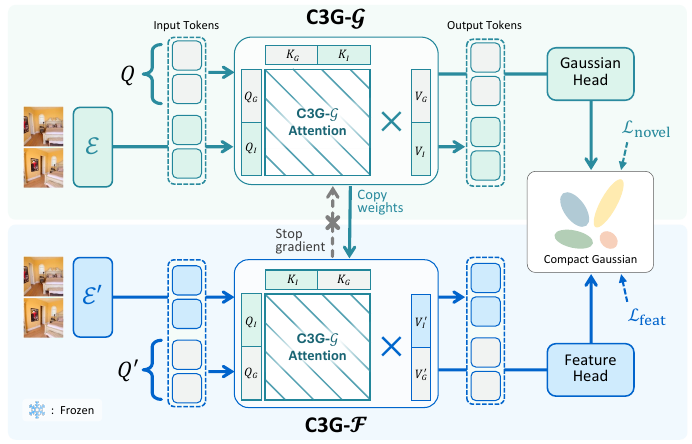}
    \caption{\textbf{\oursF\ training scheme.} We leverage the learned attention patterns from the Gaussian decoder~\mbox{\ours} (\textbf{top}) to efficiently learn a 3D feature decoder \oursF (\textbf{bottom}) for feature lifting.  We initialize \oursF\ by copying \ours's architecture and copy the attention weights from \ours, using learnable feature queries $\mathbf{Q}'$ and features $\mathcal{E}'$ from any desired encoder. Only the value projections $V'$ are trainable, enabling efficient training with $\mathcal{L}_\text{feat}$.}
    \label{fig:arch_mini}
    \vspace{-15pt}
\end{figure}
\label{subsec:feature_lifting}
Building upon the emergent property within the self-attention maps of \ours~(\textbf{\S~\ref{subsec:analysis}}), we present a simple yet effective approach for lifting arbitrary 2D features into view-invariant 3D features, dubbed \oursF.

\paragrapht{Challenges in existing feature lifting approaches.}
Previous methods~\cite{zeid2025dino,marrie2025ludvig, cheng2024occam} for lifting 2D features to 3D face two fundamental challenges: (1) \textit{Correspondence identification}: for each 2D patch whose features need to be lifted, they must identify which 3D Gaussians contribute to rendering that pixel location, often requiring computationally expensive backward mapping operations~\cite{zeid2025dino, lee2025cf3, cheng2024occam, marrie2025ludvig, jun2025dr}. (2) \textit{Multi-view inconsistency}: since image encoders extract features independently for each view, patches corresponding to the same 3D region can produce different feature representations across views, requiring additional aggregation schemes~\cite{fan2024large, burgess2006spatial}.

\paragrapht{View-invariant feature decoder.}
Surprisingly, we find that the learned self-attention patterns of \ours\ can be used to sidestep both challenges elegantly. For the correspondence identification, each learnable token shows high attention weights to the regions in each image where the 3D Gaussians are projected, removing the need for expensive backward mapping. For multi-view inconsistency, we directly use attention weights as interpolation weights to aggregate inconsistent features, instead of heuristically defining them as in previous works~\cite{cheng2024occam}.

Building on this insight, we introduce an efficient method to initialize a new \textit{view-invariant feature decoder} \oursF\, which leverages the geometric understanding already learned by our \ours~(Fig.~\ref{fig:arch_mini}). The input of \oursF\ is a set of features $\mathbf{F}_v' \in \mathbb{R}^{h \times w \times d'}$ extracted from the same set of input images $\{I_v\}_{v=1}^V$, but with a different visual encoder $\mathcal{E}'(\cdot)$ the user wants to lift to 3D. To consider different feature dimension sizes between $\mathcal{E}(\cdot)$ and $\mathcal{E}'(\cdot)$, we also initialize new learnable feature tokens $\mathbf{Q}' \in \mathbb{R}^{N \times d'}$. 

To effectively utilize the learned knowledge of the attention weights within our Gaussian decoder, we initialize \oursF\ by copying \ours's architecture and parameters, but only allow value projections inside the attention operation to be trainable while \textit{freezing the learned attention weights}. This ensures each feature token attends to the same multi-view regions as its corresponding Gaussian token, effectively reusing learned emergent correspondences within \ours\ for feature aggregation. The refined tokens $\bar{\mathbf{Q}}'$ pass through an MLP head to produce multi-view aggregated features $\mathbf{F}''_i \in \mathbb{R}^{d'}$ for each Gaussian $\mathbf{G}_i$.

The multi-view aggregated features $\mathbf{F}''_i$ are then attached to their corresponding Gaussians as additional attributes, enabling rendering of novel view feature maps via the same alpha-blending process. We train \oursF\ by minimizing the feature similarity loss with the ground truth feature and rendered feature at the target image $I_t$ at target pose $\pi_t$ as
\begin{equation}
\mathcal{L}_\text{feat} = 1 - \cos(\hat{\mathbf{F}}_t/\|\hat{\mathbf{F}}_t\|,
\mathbf{F}'_t/\|\mathbf{F}'_t\|
)\end{equation}
where $\hat{\mathbf{F}}_t$ is the rendered feature map and $\mathbf{F}'_t$ is the ground truth feature, $\cos(\cdot,\cdot)$ indicates the cosine similarity operation, and $\|\cdot\|$ is the L2-norm operation.
\section{Experiments}
\label{sec:experiment}

\subsection{Implementation details}
Here, we specify the architectural details of \ours. For the visual encoder $\mathcal{E}(\cdot)$, we adopt pretrained VGGT as default. We set $N=2048$ learnable query tokens and $L=2$ transformer layers. Following 3DGS~\cite{kerbl20233d}, we use default Gaussian attributes except setting spherical harmonics degree to 0, which we find to stabilize training with compact Gaussians by modeling only RGB color without view-directional biases. For training, we use $224 \times 224$ resolution inputs with photometric loss weights $\lambda_\text{MSE}=1$ and $\lambda_\text{LPIPS}=0.05$. We employ AdamW optimizer~\cite{loshchilov2017decoupled} with learning rates of 1e-4 for both decoders and 1e-6 for the visual encoder, using cosine annealing (minimum ratio 0.1). The model trains for 450K steps with batch size 8 per GPU across 8 NVIDIA H100 GPUs. For progressive low-pass filtering~\cite{jung2024relaxing}, we decrease $s$ from 10 to 0.3 over the first 4K steps with decay ratio 1/3 every 1K steps.

For feature lifting, we use LSeg~\cite{li2022language} and MaskCLIP~\cite{zhou2022extract} features for 3D scene understanding, and VGGT tracking features~\cite{wang2025vggt}, DINOv2~\cite{oquab2023dinov2}, and DINOv3~\cite{simeoni2025dinov3} to demonstrate \oursF's effectiveness as a view-invariant feature decoder by evaluating in two-view correspondence evaluations following Probe3D~\cite{el2024probing}. As described in \textbf{\S~\ref{subsec:feature_lifting}}, we initialize \oursF\ from \ours\ and train only value projections for 1K steps, simultaneously training both decoders.

\begin{table*}[t]
\centering
\caption{\textbf{Comparison of novel view synthesis with multi-view input images on RealEstate10K~\cite{zhou2018stereo}.} Our method generates fewer Gaussians while achieving competitive or superior quality. TTO denotes that test-time optimization is applied to the Gaussians. }
\label{tab:multiview_nvs}
\vspace{-5pt}
\resizebox{\linewidth}{!}{
\begin{tabular}{l|cccc|cccc|cccc}
\toprule
\multirow{2}{*}{Methods} & \multicolumn{4}{c|}{12 view} & \multicolumn{4}{c|}{24 view} & \multicolumn{4}{c}{36 view} \\
& PSNR$\uparrow$ & SSIM$\uparrow$ & LPIPS$\downarrow$ & \#G$\downarrow$ & PSNR$\uparrow$ & SSIM$\uparrow$ & LPIPS$\downarrow$ & \#G$\downarrow$ & PSNR$\uparrow$ & SSIM$\uparrow$ & LPIPS$\downarrow$ & \#G$\downarrow$ \\
\midrule
AnySplat~\cite{jiang2025anysplat} & 23.057& 0.807& 0.215& 1,500K& 24.105& 0.838& 0.198 & 2,636K& 24.196& 0.842& 0.192& 3,309K\\ 
AnySplat w/ TTO~\cite{jiang2025anysplat} & 26.726& 0.881& 0.189& 1,500K& 27.471& 0.898& 0.180 & 2,636K& 27.412& 0.899& 0.181& 3,309K\\ 
VGGT+NoPo~\cite{wang2025vggt, ye2024no} & 21.262& 0.667& 0.200& 602K& 21.244& 0.664& 0.200& 1,204K& 21.190& 0.663& 0.200& 1,806K\\ 
VGGT+NoPo w/ TTO~\cite{wang2025vggt, ye2024no} & 28.540& \textbf{0.898}& \textbf{0.131}& 602K& 28.463& 0.902& \textbf{0.135}& 1,204K& 28.100& 0.898& 0.145& 1,806K\\ 

\midrule
\textbf{C3G (Ours)} & 23.612& 0.740& 0.203& \textbf{2K}& 23.797& 0.747& 0.198 & \textbf{2K}& 23.812& 0.747& 0.199& \textbf{2K}\\
\textbf{C3G (Ours)} w/ TTO & \textbf{28.552}& 0.890& 0.155& 28K& \textbf{29.987}& \textbf{0.916}& 0.136& 27K & \textbf{30.250}& \textbf{0.921}& \textbf{0.133}& 26K\\
\bottomrule
\end{tabular}}
\end{table*}
\begin{figure}[t]
    \centering
    \includegraphics[width=\linewidth]{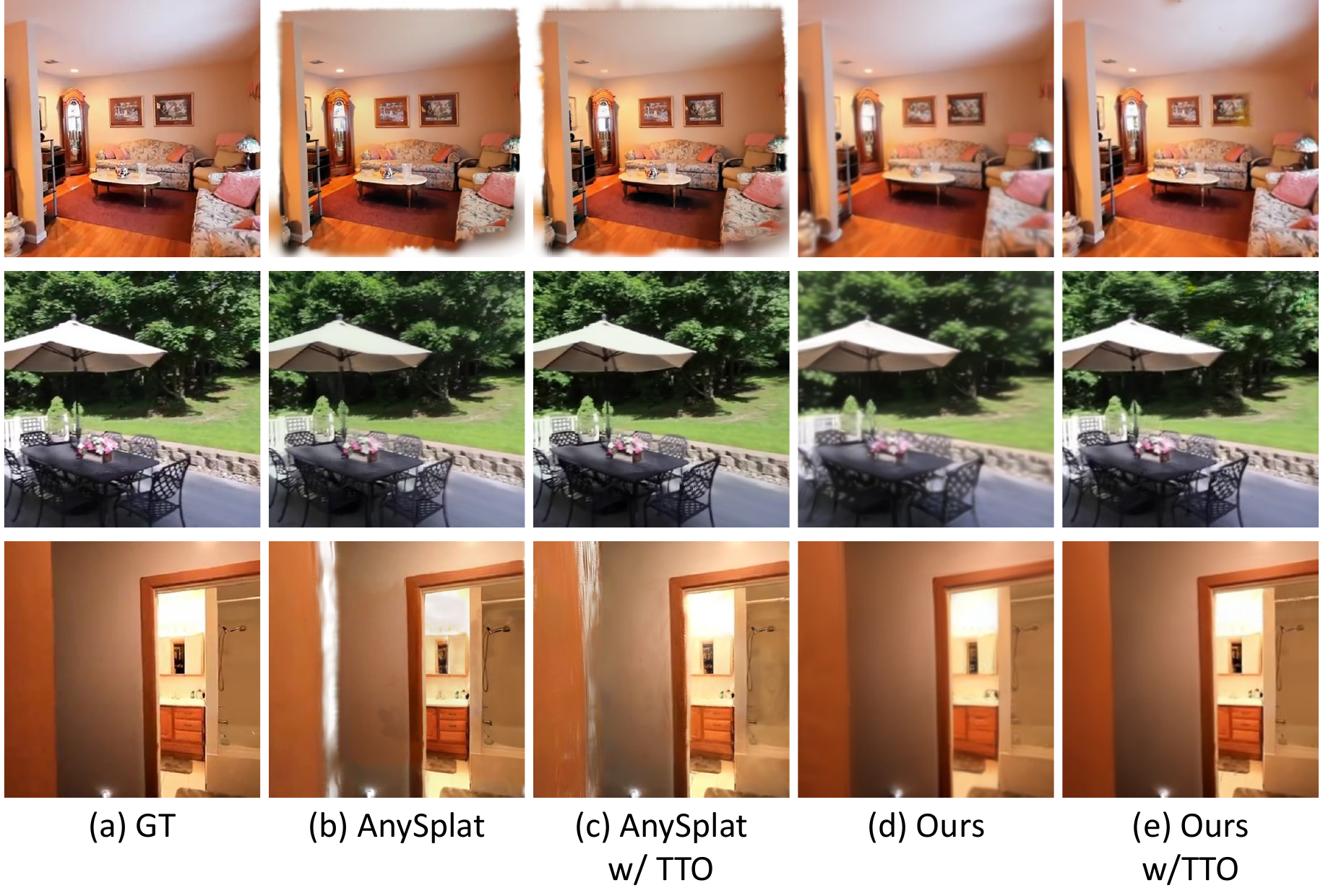}
    \vspace{-20pt}
    \caption{\textbf{Qualitative results of novel view synthesis on RealEstate10K~\cite{zhou2018stereo}.}
Given multi-view input images, our method produces the highest-quality renderings, both with and without test-time Gaussian optimization. TTO denotes that test-time optimization is applied to the Gaussians.
} \vspace{-15pt}
    \label{fig:nvs_main}
\end{figure}

\begin{table*}[t]
\centering
\caption{\textbf{Comparison of 3D scene understanding on ScanNet~\cite{dai2017scannet}.} We lift LSeg~\cite{li2022language} and MaskCLIP~\cite{zhou2022extract} features from two input views and evaluate open-vocabulary segmentation on target views. Our method generates fewer Gaussians while outperforming feed-forward and per-scene optimization methods trained with substantially more posed inputs.  $*$: Features directly extracted from target view images.}
\vspace{-10pt}
\label{tab:scannet_results}
\resizebox{\textwidth}{!}{
\begin{tabular}{lcc|ccccc|ccccc|cc}
\toprule
\multicolumn{15}{c}{Target View}\\
\midrule
\multirow{2}{*}{Methods} & \multirow{2}{*}{\shortstack{Feed\\Forward}} & \multirow{2}{*}{\shortstack{Input\\Pose}} & \multicolumn{5}{c|}{LSeg~\cite{li2022language}}& \multicolumn{5}{c|}{MaskCLIP~\cite{zhou2022extract}} & & \\
& & & mIoU$\uparrow$ & Acc$\uparrow$ & PSNR$\uparrow$ & SSIM$\uparrow$ & LPIPS$\downarrow$ & mIoU$\uparrow$ & Acc$\uparrow$ & PSNR$\uparrow$ & SSIM$\uparrow$ & LPIPS$\downarrow$ & \#G$\downarrow$ & Memories$\downarrow$ \\
\midrule
LSeg / MaskCLIP$^*$~\cite{li2022language, zhou2022extract} & \ding{55} & - & 0.506& 0.797& - & - & - & 0.341& 0.667& - & - & - & - & - \\
Feature-3DGS~\cite{zhou2024feature} & \ding{55} & \ding{51} & 0.379& 0.644& 19.83& 0.684& 0.357& 0.353& 0.663& 17.47& 0.612& 0.420& 1,185K& 845.2MB\\ 
CF$^3$~\cite{lee2025cf3} & \ding{55} & \ding{51} & 0.376& 0.657& 20.04& 0.691& 0.359& 0.336& 0.634& 20.14& 0.695& 0.354& 53K& 38.4MB\\
\midrule
LSM~\cite{fan2024large} & \ding{51} & \ding{55} & 0.503& \textbf{0.793}& 23.32& 0.767& \textbf{0.250}& 0.286& 0.505& 22.87& 0.737& \textbf{0.286}& 131K& 61.5MB\\
\textbf{C3G (Ours)} & \ding{51} & \ding{55} & \textbf{0.513}& 0.783& \textbf{23.89}& \textbf{0.770}& 0.285& \textbf{0.369}& \textbf{0.675}& \textbf{23.75}& \textbf{0.763}& 0.290& \textbf{2K} & \textbf{4.1MB}\\
\bottomrule
\toprule
\multicolumn{15}{c}{Source View}\\
\midrule
\multirow{2}{*}{Methods} & \multirow{2}{*}{\shortstack{Feed\\Forward}} & \multirow{2}{*}{\shortstack{Input\\Pose}} & \multicolumn{5}{c|}{LSeg~\cite{li2022language}}& \multicolumn{5}{c|}{MaskCLIP~\cite{zhou2022extract}} & & \\
& & & mIoU$\uparrow$ & Acc$\uparrow$ & PSNR$\uparrow$ & SSIM$\uparrow$ & LPIPS$\downarrow$ & mIoU$\uparrow$ & Acc$\uparrow$ & PSNR$\uparrow$ & SSIM$\uparrow$ & LPIPS$\downarrow$ & \#G$\downarrow$ & Memories$\downarrow$  \\
\midrule
LSeg / MaskCLIP$^*$~\cite{li2022language, zhou2022extract} & \ding{55} & - & 0.521& 0.820& -& -& -& 0.344& 0.665& -& -& -& - & -
\\
Feature-3DGS~\cite{zhou2024feature} & \ding{55} & \ding{51} & 0.392& 0.655& 21.73& 0.757& 0.314& 0.353& 0.674& 22.25& 0.777& 0.308& 1,185K& 845.2MB
\\ 
CF$^3$~\cite{lee2025cf3} & \ding{55} & \ding{51} & 0.390& 0.668& 22.99& 0.804& 0.272& 0.342& 0.642& 23.16& 0.812& 0.265& 53K& 38.4MB
\\
\midrule
LSM~\cite{fan2024large} & \ding{51} & \ding{55} & 0.511& 0.798& \textbf{25.44}& \textbf{0.811}& \textbf{0.214}& 0.251& 0.516& \textbf{25.01}& \textbf{0.824}& \textbf{0.230}& 131K& 61.5MB
\\
\textbf{C3G (Ours)} & \ding{51} & \ding{55} & \textbf{0.542}& \textbf{0.803}& 23.92& 0.766& 0.278& \textbf{0.361}& \textbf{0.668}& 23.39& 0.759& 0.284& \textbf{2K} & \textbf{4.1MB}\\
\bottomrule

\end{tabular}}
\vspace{-10pt}
\end{table*}
\begin{table*}[t]
\centering
\caption{\textbf{Comparison of 3D scene understanding on Replica~\cite{straub2019replica}.} We lift LSeg~\cite{li2022language} and MaskCLIP~\cite{zhou2022extract} features from two input views and evaluate open-vocabulary segmentation on target views. Our method generates fewer Gaussians while outperforming feed-forward methods and achieving comparable results to per-scene optimization methods trained with substantially more posed inputs.  $*$: Features directly extracted from target view images.}
\vspace{-10pt}
\label{tab:replica_results}
\resizebox{\textwidth}{!}{
\begin{tabular}{lcc|ccccc|ccccc|cc}
\toprule
\multicolumn{15}{c}{Target View}\\
\midrule
\multirow{2}{*}{Methods} & \multirow{2}{*}{\shortstack{Feed\\Forward}} & \multirow{2}{*}{\shortstack{Input\\Pose}} & \multicolumn{5}{c|}{LSeg~\cite{li2022language}}& \multicolumn{5}{c|}{MaskCLIP~\cite{zhou2022extract}} & & \\
& & & mIoU$\uparrow$ & Acc$\uparrow$ & PSNR$\uparrow$ & SSIM$\uparrow$ & LPIPS$\downarrow$ & mIoU$\uparrow$ & Acc$\uparrow$ & PSNR$\uparrow$ & SSIM$\uparrow$ & LPIPS$\downarrow$ & \#G$\downarrow$ & Memories$\downarrow$ \\
\midrule
LSeg / MaskCLIP$^*$~\cite{li2022language, zhou2022extract} & \ding{55} & - & 0.618& 0.887& -& -& -& 0.412& 0.668& -& -& -& -& - \\
Feature-3DGS~\cite{zhou2024feature} & \ding{55} & \ding{51} & 0.730& 0.936& 35.70& 0.972& 0.044& 0.421& 0.686& 35.90& 0.972& 0.045& 199K& 141.8MB\\ 
CF$^3$~\cite{lee2025cf3} & \ding{55} & \ding{51} & 0.663& 0.918& 27.49& 0.906& 0.132& 0.380& 0.654& 27.49& 0.906& 0.132& 10K& 7.1MB\\
\midrule
LSM~\cite{fan2024large} & \ding{51} & \ding{55} & 0.600& 0.823& 21.86& 0.753& 0.213& 0.241& 0.411& 17.01& 0.637& 0.377& 
131K& 61.5MB\\
\textbf{C3G (Ours)} & \ding{51} & \ding{55} & \textbf{0.630}& \textbf{0.893}& \textbf{25.43}& \textbf{0.818}& \textbf{0.173}& \textbf{0.416}& \textbf{0.692}& \textbf{25.00}& \textbf{0.809}& \textbf{0.182}& \textbf{2K}& \textbf{4.1MB}\\
\bottomrule
\toprule
\multicolumn{15}{c}{Source View}\\
\midrule
\multirow{2}{*}{Methods} & \multirow{2}{*}{\shortstack{Feed\\Forward}} & \multirow{2}{*}{\shortstack{Input\\Pose}} & \multicolumn{5}{c|}{LSeg~\cite{li2022language}}& \multicolumn{5}{c|}{MaskCLIP~\cite{zhou2022extract}} & & \\
& & & mIoU$\uparrow$ & Acc$\uparrow$ & PSNR$\uparrow$ & SSIM$\uparrow$ & LPIPS$\downarrow$ & mIoU$\uparrow$ & Acc$\uparrow$ & PSNR$\uparrow$ & SSIM$\uparrow$ & LPIPS$\downarrow$ & \#G$\downarrow$ & Memories$\downarrow$ \\
\midrule
LSeg / MaskCLIP$^*$~\cite{li2022language, zhou2022extract} & \ding{55} & - & 0.647& 0.896& -& -& -& 0.414& 0.674& -& -& -& -& -\\
Feature-3DGS~\cite{zhou2024feature} & \ding{55} & \ding{51} & 0.729& 0.930& 36.46& 0.975& 0.043& 0.416& 0.680& 36.63& 0.975& 0.043& 199K& 141.8MB
\\ 
CF$^3$~\cite{lee2025cf3} & \ding{55} & \ding{51} & 0.664& 0.916& 27.95& 0.913& 0.127& 0.375& 0.649& 27.95& 0.913& 0.127& 10K& 7.1MB
\\
\midrule
LSM~\cite{fan2024large} & \ding{51} & \ding{55} & 0.600& 0.823& 19.27& 0.760& 0.230& 0.241& 0.439& 17.53& 0.670& 0.377& 131K& 61.5MB
\\
\textbf{C3G (Ours)} & \ding{51} & \ding{55} & \textbf{0.649}& \textbf{0.894}& \textbf{25.35}& \textbf{0.815}& \textbf{0.177}& \textbf{0.421}& \textbf{0.695}& \textbf{25.07}& \textbf{0.811}& \textbf{0.185}& \textbf{2K}& \textbf{4.1MB}\\
\bottomrule

\end{tabular}}
\vspace{-5pt}
\end{table*}

\begin{figure*}[t]
    \centering
    \includegraphics[width=\linewidth]{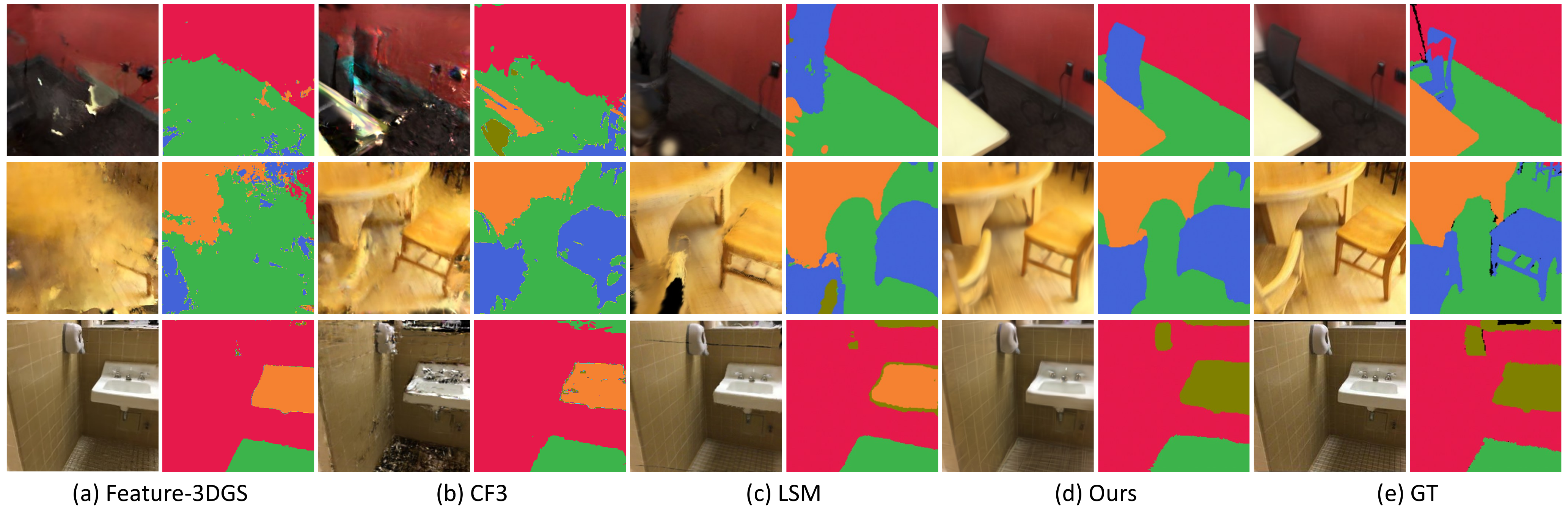}
    \vspace{-20pt}
    \caption{\textbf{Qualitative results of 3D scene understanding on ScanNet~\cite{dai2017scannet}.} We conduct qualitative comparison for 3D scene understanding via novel view synthesis and open-vocabulary segmentation. When compared to both per-scene optimization~((a), (b)) and feed-forward ~((c), (d)) methods, ours show the most high-fidelity renderings and accurate segmentation maps compared to the ground-truth.} 
    \label{fig:understanding_main}\vspace{-10pt}
\end{figure*}

\begin{table}[t]
\centering
\caption{\textbf{Correspondence estimation on ScanNet~\cite{dai2017scannet}.} We evaluate PCK@10px across two images captured from different angles. Our feature aggregation significantly improves correspondence accuracy across the VGGT-Tracking, DINOv2, and DINOv3.}
\vspace{-5pt}
\resizebox{\linewidth}{!}{
\begin{tabular}{l|cccc|c}
\toprule
Methods & $\theta_{0}^{15}$ & $\theta_{15}^{30}$ & $\theta_{30}^{60}$ & $\theta_{60}^{180}$ & Avg. \\
\midrule
VGGT-Tracking~\cite{wang2025vggt}& 53.6& 45.5& 27.6& 9.3& 34.0\\
VGGT-Tracking~\cite{wang2025vggt} + AnyUp~\cite{wimmer2025anyup}& 58.7& 50.4& 35.7& 11.8& 39.2\\
VGGT-Tracking~\cite{wang2025vggt} + \textbf{C3G (Ours)}& \textbf{93.5}& \textbf{88.1}& \textbf{70.4}& \textbf{20.3}& \textbf{68.1}\\
\midrule
DINOv2-Base~\cite{oquab2023dinov2} & 36.9& 29.2& 19.3& 9.9& 23.8\\
DINOv2-Base~\cite{oquab2023dinov2} + FiT3D~\cite{yue2024improving} & 40.9 & 32.1 & 22.0 & 13.9 & 27.2 \\
DINOv2-Base~\cite{oquab2023dinov2} + AnyUp~\cite{wimmer2025anyup}& 36.2 & 28.6 & 19.7 & 10.9 & 23.9\\
DINOv2-Base~\cite{oquab2023dinov2} + \textbf{C3G (Ours)} & \textbf{92.9} & \textbf{88.4} & \textbf{70.5} & \textbf{22.2} &  \textbf{68.5}\\
\midrule
DINOv2-Large~\cite{oquab2023dinov2} & 36.6& 24.0& 19.0& 12.7& 23.1\\
DINOv2-Large~\cite{oquab2023dinov2} + AnyUp~\cite{wimmer2025anyup}& 39.4& 25.5& 19.1& 14.0& 24.5\\
DINOv2-Large~\cite{oquab2023dinov2} + \textbf{C3G (Ours)} & \textbf{94.2}& \textbf{89.0}& \textbf{70.5}& \textbf{21.0}& \textbf{68.7}\\
\midrule
DINOv3-Large~\cite{simeoni2025dinov3} &  54.9& 44.3& 32.1& 19.3& 37.7\\
DINOv3-Large~\cite{simeoni2025dinov3} + AnyUp~\cite{wimmer2025anyup}& 48.4& 34.8& 27.1& 16.8& 31.8\\
DINOv3-Large~\cite{simeoni2025dinov3} + \textbf{C3G (Ours)} & \textbf{94.2}& \textbf{89.1}& \textbf{70.8}& \textbf{20.9}& \textbf{68.8}\\
\bottomrule
\end{tabular}}
\label{tab:correspondence}
\vspace{-10pt}
\end{table}
\begin{figure}[t]
    \centering
    \includegraphics[width=\linewidth]{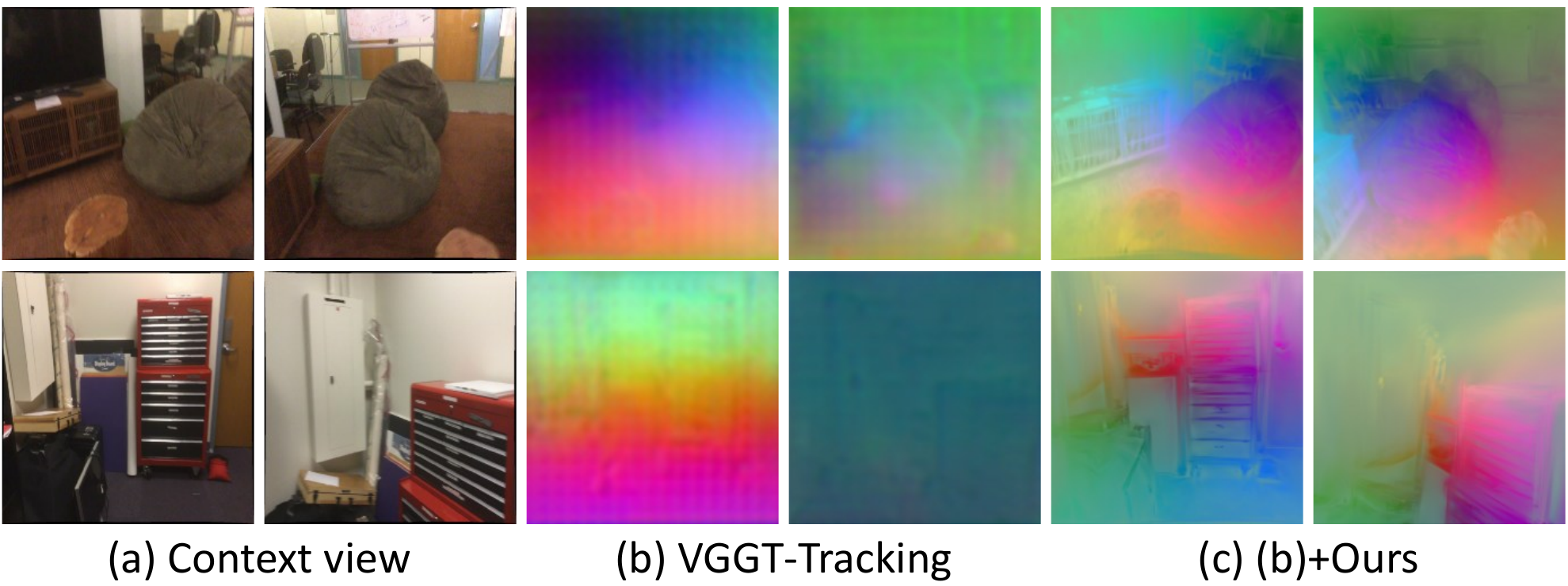}
    \vspace{-15pt}
    \caption{\textbf{PCA visualization of multi-view features on ScanNet~\cite{dai2017scannet}.} We visualize the PCA results of encoded multi-view features. Our method improves multi-view consistency compared to the original visual features~\cite{wang2025vggt}.} 
    \label{fig:correspondence}\vspace{-10pt}
\end{figure}

\begin{figure}[t]
    \centering
    \includegraphics[width=\linewidth]{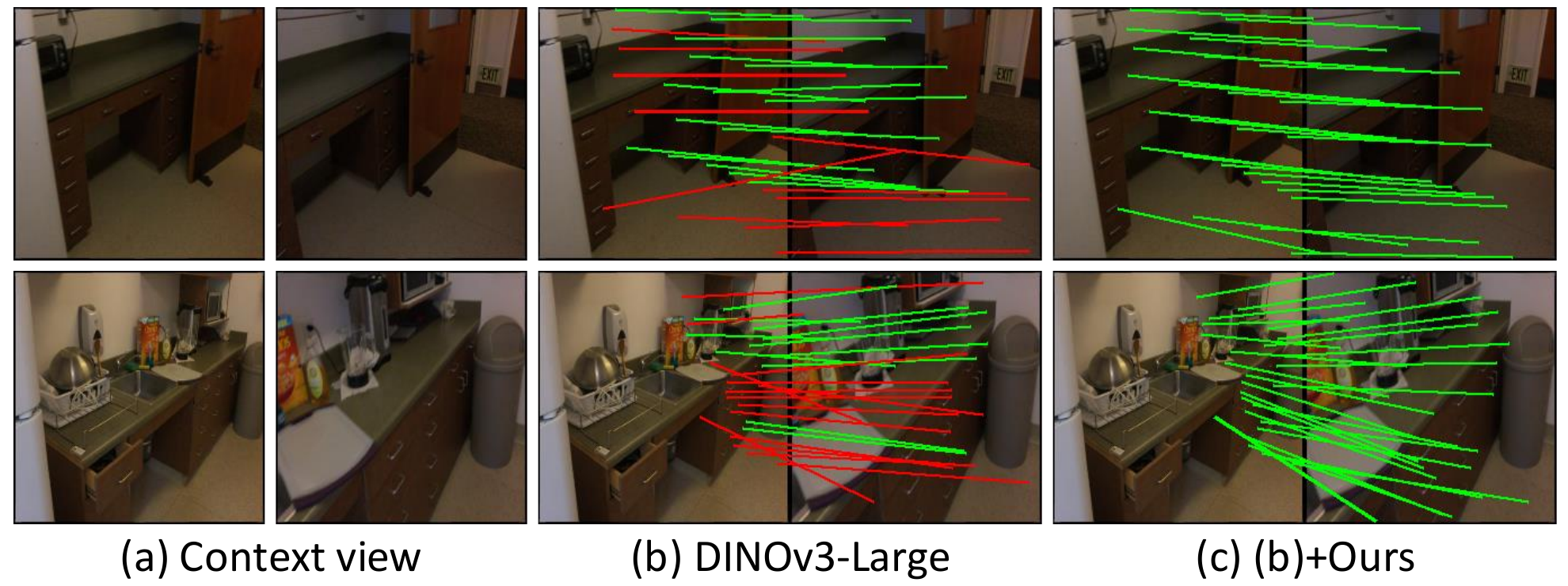}
    \vspace{-15pt}
    \caption{\textbf{Visualization of estimated correspondence of multi-view features on ScanNet~\cite{dai2017scannet}.} We visualize the estimated correspondence of encoded multi-view features. The \textcolor[HTML]{00FF00}{green} line indicates inliers, and the \textcolor[HTML]{FF0000}{red} line indicates outliers. Our method significantly improves multi-view correspondence compared to the original visual features~\cite{simeoni2025dinov3}.} 
    \label{fig:correspondence_viz}\vspace{-10pt}
\end{figure}

\begin{table}[t]
\centering
\caption{\textbf{Ablation studies for \ours\ on RealEstate10K~\cite{zhou2018stereo}.}}
\label{tab:stage1}
\vspace{-5pt}
\resizebox{0.7\linewidth}{!}{
\begin{tabular}{cc|ccc}
\toprule
\multirow{2}{*}{\shortstack{Lowpass\\Filter}} & \multirow{2}{*}{\shortstack{Unfreeze\\Backbone}} & \multicolumn{3}{c}{Average} \\
&&PSNR$\uparrow$ & SSIM$\uparrow$ & LPIPS$\downarrow$ \\
\midrule
\ding{55} & \ding{51} &N/A&N/A& N/A\\
\ding{51} & \ding{55} &18.441&0.553& 0.408\\
\ding{51} & \ding{51} & \textbf{22.387} & \textbf{0.713} & \textbf{0.259} \\
\bottomrule
\end{tabular}}

\end{table}
\begin{table}[t]
\centering
\caption{\textbf{Ablation studies on the number of Gaussian.}}
\vspace{-5pt}
\resizebox{0.5\linewidth}{!}{
\begin{tabular}{c|ccc}
\toprule
\#G & PSNR$\uparrow$ & SSIM$\uparrow$ & LPIPS$\downarrow$\\
\midrule
256 &19.713&0.589& 0.425\\
512 &20.223&0.607& 0.378\\
1024 &20.559&0.619& 0.338\\
2048 &\textbf{20.625}&\textbf{0.623}&\textbf{0.321} \\
4096 &19.012&0.568& 0.450\\
\bottomrule
\end{tabular}}

\label{tab:num_gaussian}
\end{table}

\subsection{Novel view synthesis}
In this section, we evaluate novel view synthesis performance on the RealEstate10k dataset~\cite{zhou2018stereo} with 12, 24, and 36 input images following AnySplat's~\cite{jiang2025anysplat} protocol, where multiple images are used to estimate 3D Gaussians and render a target view. We compare AnySplat~\cite{jiang2025anysplat} as it was trained to support multi-view inputs. As we utilize VGGT as our visual encoder, we additionally train and compare with VGGT+NoPo, which trains a NoPoSplat-like model for per-pixel 3DGS estimation but replaces the backbone from MASt3R~\cite{leroy2024grounding} with VGGT~\cite{wang2025vggt}.  As shown in Tab.~\ref{tab:multiview_nvs}, while VGGT+NoPo generates per-pixel Gaussians without pruning and AnySplat reduces Gaussians via voxel merging, our method achieves comparable performance with significantly fewer Gaussians. In Fig.~\ref{fig:nvs_main}, we provide qualitative results for novel view synthesis, showing that our method is competitive with AnySplat and generates fewer artifacts. We also perform short test-time optimization~(TTO) following 3DGS~\cite{kerbl20233d}, where our method substantially outperforms both models. Note that VGGT+NoPo's performance degrades as the number of input images increases due to the accumulation of alignment error. For AnySplat and VGGT+NoPo with TTO, we follow the default test-time optimization of~\cite{jiang2025anysplat}, which disable Gaussian densification, as standard 3DGS optimization results in $\texttt{Out-Of-Memory}$ errors due to the already excessive number of Gaussians.

\subsection{3D scene understanding}
Following previous approaches~\cite{fan2024large, burgess2006spatial}, we evaluate 3D scene understanding on ScanNet~\cite{dai2017scannet} and Replica~\cite{straub2019replica}. To enable open-vocabulary segmentation~\cite{cho2024cat,shin2024towards} at novel viewpoints, we lift language-aligned features (LSeg~\cite{li2022language}, MaskCLIP~\cite{zhou2022extract}) extracted from two input views and evaluate with rendered features at target poses. We compare with feed-forward approaches~\cite{fan2024large} and per-scene optimization methods~\cite{zhou2024feature, lee2025cf3}. Note that feed-forward methods use two input images, while optimization-based methods use all scene images for training. As shown in Tab.~\ref{tab:scannet_results} and Tab.~\ref{tab:replica_results}, our method outperforms LSM~\cite{fan2024large} in segmentation with competitive or better reconstruction quality. Despite using far fewer input images, we outperform optimization-based methods on ScanNet and show comparable performance on Replica. We additionally provide qualitative results in Fig~\ref{fig:understanding_main}, showing that our model achieves high rendering quality and the most accurate segmentation results. Surprisingly, our rendered features at target views obtained without accessing the target image outperform features directly extracted from target images using LSeg or MaskCLIP. This validates the effectiveness of our compact Gaussians over previous per-pixel Gaussians, and verifies that \oursF\ effectively aggregates features from input features, enabling the rendering of a multi-view aware feature map.

\subsection{Multi-view feature encoding}
In this section, we further validate the effectiveness of \mbox{\oursF} as a view-invariant feature encoder. By lifting the multi-view aggregated features from \oursF\ and re-rendering to the input view camera poses, we can achieve view-invariant features from the input views. Following Probe3D~\cite{el2024probing}, we evaluate two-view correspondence performance in ScanNet~\cite{dai2017scannet} with \mbox{PCK $@$ 10px}, where the selected two views are captured within $0{\sim}15^\circ$~($\theta_0^{15}$), $15{\sim}30^\circ$~($\theta_{15}^{30}$), $30{\sim}60^\circ$~($\theta_{30}^{60}$), and $60{\sim}180^\circ$~($\theta_{60}^{180}$). We compare the features from the visual encoder $\mathcal{E}'(\cdot)$ and the features obtained from our re-rendering process. We evaluate with three different visual encoders, including the tracking features of VGGT~\cite{wang2025vggt}, DINOv2~\cite{oquab2023dinov2}, and DINOv3~\cite{simeoni2025dinov3}. As shown in Tab.~\ref{tab:correspondence}, Fig.~\ref{fig:correspondence}, and Fig.~\ref{fig:correspondence_viz}, the aggregated features show significant performance improvement in all settings. 
We also compare our model with FiT3D~\cite{yue2024improving}, which finetunes 2D vision encoders~\cite{oquab2023dinov2, du2023learning} by lifting features to 3D Gaussians to enforce multi-view consistency. However, FiT3D uses an autoencoder architecture that introduces information loss and generates a large number of Gaussians, making effective multi-view feature aggregation challenging. In contrast, our model demonstrates significantly superior performance, validating the effectiveness of \oursF\ as a view-invariant feature decoder.

\subsection{Multi-view feature upsampling}
Since 3D Gaussians can be projected to arbitrary camera poses and intrinsics, we can render images at different resolutions. This property enables upsampled feature generation even when the visual encoder $\mathcal{E}'(\cdot)$ produces only coarse resolution features. Since our model can lift features from any visual encoder, we compare with the current state-of-the-art upsampler, AnyUp~\cite{wimmer2025anyup}. AnyUp can also upsample any features to the original image resolution, so we evaluate various feature extractors, including VGGT-Tracking~\cite{wang2025vggt}, DINOv2-Base~\cite{oquab2023dinov2}, DINOv2-Large~\cite{oquab2023dinov2}, and DINOv3-Large~\cite{fan2024large}. 

In Tab.~\ref{tab:correspondence}, we evaluate the correspondence performance of the upsampled features. While feature upsampling generally improves correspondence~\cite{leroy2024grounding, edstedt2024roma} and AnyUp effectively upsamples features to the original image resolution, feature upsampled by AnyUp fails to consistently improve correspondence performance across all baselines, and even degrades DINOv3's performance. In contrast, our model not only effectively upsamples the multi-view features but also enhances their multi-view consistency, leading to substantial improvements in correspondence performance.

\subsection{Ablation studies}
\label{subsec:ablation}
In this section, we go through the ablation studies done for \ours\ and \oursF. In Tab.~\ref{tab:stage1}, we validate our core training components. Without progressive low-pass filter control, 3D Gaussians fail to localize within the view frustum, causing training collapse. We also validate that freezing the visual encoder $\mathcal{E}(\cdot)$ prevents effective Gaussian generation in appropriate regions. In Tab.~\ref{tab:num_gaussian}, we investigate the optimal number of learnable tokens (Gaussians) $N$. Reconstruction performance gradually improves as Gaussians increase to 2048. However, with 4096 Gaussians, training becomes unstable, leading to degradation. We hypothesize that having larger number of Gaussians at sub-optimal positions is more prone to falling into local minima, as discussed in prior per-scene optimization methods~\cite{jung2024relaxing}. Based on these findings, we set the number of Gaussians to 2048 for all experiments. In Tab.~\ref{tab:feature_backbone}, we analyze different visual encoders $\mathcal{E}(\cdot)$ for \ours\, using VGGT~\cite{wang2025vggt} and DINOv3~\cite{simeoni2025dinov3}. Although DINOv3 lacks explicit geometric supervision, its features are effectively aggregated by learnable queries in the transformer to generate coherent 3D Gaussians. These results reveal the potential that our framework can also be learned without features with strong geometric priors, where previous per-pixel Gaussian estimation frameworks struggle to learn~\cite{ye2024no}.

\begin{table}[t]
\centering
\caption{\textbf{Ablation studies for \oursF\ on Scannet~\cite{dai2017scannet}.}}
\label{tab:ablation_stage2}
\vspace{-5pt}
\resizebox{\linewidth}{!}{
\begin{tabular}{cc|ccccc}
\toprule
\oursF & \shortstack{Autoencoder}
&mIOU$\uparrow$ & Acc.$\uparrow$& PSNR$\uparrow$ & SSIM$\uparrow$ & LPIPS$\downarrow$ \\
\midrule
\ding{55} & \ding{55} &0.193&0.413&21.77&0.706& 0.418\\
\ding{51} & \ding{51} &0.512&0.782&23.604&0.756& 0.277\\
\ding{51} & \ding{55} &\textbf{0.513}&\textbf{0.783}&\textbf{23.886}&\textbf{0.770}& \textbf{0.285}\\
\bottomrule
\end{tabular}}
\end{table}
\begin{table}[t]
\centering
\caption{\textbf{Ablation stuides on visual encoder choice.}}
\label{tab:feature_backbone}
\vspace{-5pt}
\resizebox{0.7\linewidth}{!}{
\begin{tabular}{l|ccc}
\toprule
Backbones &PSNR$\uparrow$ & SSIM$\uparrow$ & LPIPS$\downarrow$ \\
\midrule
VGGT~\cite{wang2025vggt} &\textbf{22.387}&\textbf{0.713}& \textbf{0.259}\\
DINOv3~\cite{simeoni2025dinov3} &20.292&0.631& 0.313\\
\bottomrule
\end{tabular}}
\vspace{-10pt}
\end{table}

\section{Conclusion}
\label{sec:conclusion}
We presented \ours, a framework that learns compact 3D Gaussians with learnable query tokens, which can discover geometrically meaningful regions through self-attention, resolving redundancy and high computational overhead issues from previous dense per-pixel predictions. In addition, we present \oursF, which leverages the attention weights learned in \ours\ to effectively decode multi-view consistent features. By combining both \ours\ and \oursF\, we achieve competitive performance in novel view synthesis and outperform previous works in 3D scene understanding while being significantly more efficient in memory and rendering speeds. We believe that our approach opens up new directions for feed-forward reconstruction and scene understanding, mitigating the need for per-pixel estimations.

\section*{Acknowledgment}
This research was supported by Institute of Information \& communications Technology Planning \& Evaluation (IITP) grant funded by the Korea government (MSIT) (RS-2019-II190075, RS-2024-00509279, RS-2025-II212068, RS-2023-00227592, RS-2025-02214479, RS-2024-00457882, RS-2025-25441838, RS-2025-25441838, RS-2025-02214479, RS-2025-02217259) and the Culture, Sports, and Tourism R\&D Program through the Korea Creative Content Agency grant funded by the Ministry of Culture, Sports and Tourism (RS-2024-00345025, RS-2024-00333068, RS-2023-00222280, RS-2023-00266509), and National Research Foundation of Korea (RS-2024-00346597). This research was supported by the ETH AI Center through an ETH AI Center postdoctoral fellowship to Sunghwan Hong.

\clearpage
\appendix

\section*{\Large Appendix}

This appendix provides additional details on implementation, evaluation metrics, and qualitative results that could not be included in the main document due to space constraints. The document is organized as follows:

\begin{itemize}
    \item \textbf{Section~\ref{supsec:multidata}} presents extended experiments in which \ours\ is trained on multiple diverse datasets to improve its scalability and generalizability.

    \item \textbf{Section~\ref{sec:implementation}} provides additional implementation details regarding the architecture of the Gaussian Decoder and the training details for Any-Feature 3D Lifting.
    \item \textbf{Section~\ref{sec:settings}} details the experimental settings, including dataset descriptions, baseline configurations, and evaluation protocols for efficient novel view synthesis, 3D scene understanding, and multi-view feature encoding.
    \item \textbf{Section~\ref{sec:ablation_discussion}} presents additional ablation studies comparing compact reconstruction strategies and analyzing components of \oursF, followed by discussions on the autoencoder design, FPS analysis, and limitations.
    \item \textbf{Section~\ref{sec:additional_quant}} provides additional quantitative experimental results, including latent decoding or two-view novel view synthesis.
    \item \textbf{Section~\ref{sec:additional_qual}} presents additional qualitative results, including visualizations of attention maps, 3D scene understanding, feature PCA, and renderings from 24-view inputs.
\end{itemize}

\section{Multi Dataset Training}
\label{supsec:multidata}
We extend C3G to C3G++ to improve its scalability and generalizability by incorporating diverse training datasets, increasing the number of input views, and scaling up the number of Gaussians. However, since each dataset exhibits different camera parameters and scene scales, naively combining them makes training C3G unstable and difficult to converge. To address this, we first describe the architectural modifications made to accommodate these variations (\S~\ref{supsubsec:method}), followed by implementation details covering dataset configurations and training procedures (\S~\ref{supsubsec:imde}). Finally, we present experimental results demonstrating the effectiveness of this extension (\S~\ref{supsubsec:exp}).

\begin{table*}[t]
\centering
\caption{\textbf{Comparison of novel view synthesis with multi-view input images on DL3DV~\cite{ling2024dl3dv}.} Our method generates fewer Gaussians while achieving competitive or superior quality.}
\label{suptab:multiview_nvs_dl3dv}
\vspace{-10pt}
\resizebox{\linewidth}{!}{
\begin{tabular}{l|cccc|cccc|cccc}
\toprule
\multirow{2}{*}{Methods} & \multicolumn{4}{c|}{12 view} & \multicolumn{4}{c|}{24 view} & \multicolumn{4}{c}{36 view} \\
& PSNR$\uparrow$ & SSIM$\uparrow$ & LPIPS$\downarrow$ & \#G$\downarrow$ & PSNR$\uparrow$ & SSIM$\uparrow$ & LPIPS$\downarrow$ & \#G$\downarrow$ & PSNR$\uparrow$ & SSIM$\uparrow$ & LPIPS$\downarrow$ & \#G$\downarrow$ \\
\midrule
AnySplat~\cite{jiang2025anysplat} & 17.004& 0.439& 0.431& 570K& 17.263& 0.445& 0.439& 1,140K& 18.286& 0.525& 0.286& 1,576K\\ 
VGGT+NoPo~\cite{wang2025vggt, ye2024no} & 11.714& 0.320& 0.550& 602K& 10.285& 0.262& 0.607& 1,204K& 9.65 & 0.236 & 0.628 & 1,806K\\ 

\midrule
\textbf{C3G (Ours)} & 18.768& 0.505& 0.408& 2K& 18.843& 0.511& 0.404& 2K& 18.675& 0.504& 0.412& 2K\\
\textbf{C3G++ (Ours)} & 19.368& 0.512& 0.431& 32K& 19.570& 0.521& 0.425& 32K& 19.736& 0.523& 0.424& 32K\\
\bottomrule
\end{tabular}}
\vspace{-5pt}
\end{table*}

\subsection{Methodology}
\label{supsubsec:method}
\paragrapht{Architecture modification.}
To support training on diverse datasets with varying numbers of input views and an increased number of Gaussians, we introduce minimal modifications to the original \ours\ architecture. To fully exploit the pretrained weights of \ours, which were already well-trained on the RealEstate10K dataset~\cite{zhou2018stereo}, we keep the architectural changes as minimal as possible to preserve the original structure.

First, to handle differences in the intrinsics of the camera and the scale of the scene between training datasets, we incorporate the intrinsic embeddings as follows~\cite{ye2024no, jiang2025anysplat}. Specifically, we additionally take camera intrinsics as input and pass them through a projection layer consisting of a 2-layer MLP initialized with zero weights. The projected intrinsic features are then added to the patch-embedded features within VGGT~\cite{wang2025vggt}.

Next, we increase the number of Gaussians to better cover larger scenes and capture finer details. To achieve this, we replicate the weights of the Gaussian head. In the original architecture, the Gaussian head generates one Gaussian per token. By replicating its weights $N_{\mathbf{G}}$ times, we instead produce $N_{\mathbf{G}}$ Gaussians per token, resulting in a total of $N \times N_{\mathbf{G}}$ Gaussians. Since the replicated heads share the same initialization, Gaussians originating from the same token are localized in the same spatial region, encouraging them to focus on capturing finer local details.

\paragrapht{Additional loss.}
To improve training stability, we introduce additional geometry supervision in the form of depth and normal losses, encouraging the network to develop a stronger understanding of scene geometry.
We first extract pseudo ground-truth normal maps $N_t$ and depth maps $D_t$ from the point maps produced by VGGT~\cite{wang2025vggt} for the corresponding view $I_t$. We then render the depth maps $\hat{D}_t$ and normal maps $\hat{N}_t$ using the same rasterizer as 3DGS~\cite{kerbl20233d}, replacing the color attributes with $z$-value and surface normals. For supervision, we apply scale-shift invariant loss term for depth and MSE loss for normals,
\begin{equation}
    \mathcal{L}_\text{depth} = \| (\alpha \cdot D_t + \beta) - \hat{D}_t \|,
\end{equation}
\begin{equation}
    \mathcal{L}_\text{normal} = \| N_t - \hat{N}_t \|,
\end{equation}
where $\alpha$ and $\beta$ is the scale and shift parameters obtained via least-squares alignment between rendered depth and pseudo ground-truth depth, following \cite{wang2025moge}. 

The overall loss term is followed as:
\begin{equation}
    \mathcal{L} = \mathcal{L}_{\text{novel}} + \lambda_{\text{depth}} \mathcal{L}_{\text{depth}} + \lambda_{\text{normal}} \mathcal{L}_{\text{normal}},
\end{equation}
where $\lambda_{\text{depth}}$ and $\lambda_{\text{normal}}$ denotes the loss weight of normal and depth.

\subsection{Implementation details}
\label{supsubsec:imde}
For efficient training, we initialize from the pretrained weights of \ours\, which were trained solely on the RealEstate10K dataset~\cite{zhou2018stereo}. We set the number of Gaussians per query token to $N_{\mathbf{G}} = 16$. During training, we randomly sample 2 to 24 images per scene, with a maximum of 192 images per epoch, and train at a resolution of $224 \times 224$. We use the AdamW optimizer~\cite{kingma2014adam} with a learning rate of 1e-6. The geometry loss weights are set to $\lambda_{\text{depth}}=0.01$ and $\lambda_{\text{normal}}=0.001$. We train for 150K steps in total, with all other hyperparameters following \ours.

For training, we incorporate 8 datasets spanning diverse domains. Specifically, we use RealEstate10K~\cite{zhou2018stereo}, DL3DV~\cite{ling2024dl3dv}, HyperSim~\cite{roberts2021hypersim}, WildRGBD~\cite{xia2024rgbd}, TartanAir~\cite{wang2020tartanair}, ARKitScenes~\cite{baruch2021arkitscenes}, BlendedMVS~\cite{yao2020blendedmvs}, and MapFree~\cite{arnold2022map}. These datasets collectively cover a wide range of scenarios, including indoor, outdoor, and object-centric scenes across both real and synthetic domains.

\subsection{Experiments}
\label{supsubsec:exp}
We evaluate novel-view synthesis performance on the test split of the DL3DV dataset~\cite{ling2024dl3dv}. As shown in Tab.~\ref{suptab:multiview_nvs_dl3dv}, our model outperforms the baselines, benefiting from its improved generalizability and the increased number of Gaussians, which together enable better coverage of large scenes and finer detail capture. This is particularly evident on DL3DV, which predominantly consists of outdoor scenes with large spatial extents that are difficult to represent with only 2K Gaussians. By scaling up the number of Gaussians, our model can cover a significantly larger portion of each scene, leading to improved reconstruction quality. Furthermore, the results demonstrate that our model is robust to varying numbers of input views.


\section{Additional implementation details}
\label{sec:implementation}
\subsection{Details of Gaussian decoder}
\paragrapht{Architecture details.}
We first extract feature maps $\mathbf{F}_v = \mathcal{E}(I_v)$ from the visual encoder $\mathcal{E}(\cdot)$ given $V$ input images $\{I_v\}_{v=1}^V$. For VGGT~\cite{wang2025vggt} which supports multi-view inputs, we directly obtain $V$ feature maps $\{\mathbf{F}_v\}_{v=1}^V = \mathcal{E}(\{I_v\}_{v=1}^V)$ in a single feed-forward pass. For \mbox{DINOv3~\cite{simeoni2025dinov3}}, we obtain $V$ feature maps by passing each image to the encoder independently. We then concatenate the feature maps with the learnable tokens $\mathbf{Q}$, which are randomly initialized. This concatenated representation is processed through $L$ transformer layers $\mathcal{T}_\mathcal{G}(\cdot)$, which consists of self-attention layers and MLP layers with ReLU activation functions and layer normalization~\cite{ba2016layer}:
\begin{equation}
    [\bar{\mathbf{Q}};\bar{\mathbf{F}}] = \mathcal{T}_\mathcal{G}([\mathbf{Q};\mathbf{F}]),
\end{equation}
where $[\cdot,\cdot]$ denotes concatenation of dimension axis, $\bar{\mathbf{Q}}$ denotes refined learnable tokens and $\bar{\mathbf{F}}$ denotes refined features. 
In the self-attention layer, learnable tokens and the feature tokens are each projected to query, key, and value features to process the attention calculation. The query, key and value features of learnable tokens $\mathbf{Q}$ and visual encoder features $\mathbf{F}$ is calculated as follows:
\begin{align}
Q = [Q_G;Q_I] = \mathcal{P}_Q([\mathbf{Q};\mathbf{F}]), \\ K = [K_G;K_I] = \mathcal{P}_K([\mathbf{Q};\mathbf{F}]), \\ V =[V_G;V_I] = \mathcal{P}_V([\mathbf{Q};\mathbf{F}]),
\end{align}
where $Q_G, Q_I$ denotes query features, $K_G, K_I$ denotes key features, and $V_G, V_I$ denotes value features of learnable token and visual encoder features, respectively. And $\mathcal{P}_Q$, $\mathcal{P}_K$, and $\mathcal{P}_V$ denote the projection layer of query, key, and value, respectively.

Then, the resulting output of self-attention layer in $\mathcal{T}_G$ is computed as:
\begin{equation}
    \mathsf{Attn}(Q,K,V) = \mathsf{Softmax}(\frac{QK^\intercal}{\sqrt{d_Q}})V,
\end{equation}
where $d_Q$ denotes the channel dimension of query features.

Each of the refined learnable tokens $\bar{\mathbf{Q}}_i$ are decoded as a single Gaussian $\mathbf{G}_i$ by passing the Gaussian head $H_\mathcal{G}(\cdot)$, which consists of a single linear layer:
\begin{equation}
    \mathbf{G}_i = H_\mathcal{G}(\bar{\mathbf{Q}}_i).
\end{equation}

\paragrapht{Attention visualization.}
To visualize the attention map in Fig.~3-(b), we select one Gaussian $\mathbf{G}_i$ and its corresponding learnable tokens $\mathbf{Q}_i$. We then calculate the attention weights between the query features of the learnable token $Q_{G_i}$ and the key features $K_I$ as follows:
\begin{equation}
    A_i =  Q_{G_i} \cdot K_I^\intercal, A_i \in \mathrm{R}^{N_\text{head} \times h \times w},
\end{equation}
where $N_\text{head}$ denotes number of heads. We average the $A_i$ at the head dimension, then apply min-max normalization for better visualization.

\subsection{Details of feature decoder}
\paragrapht{Architecture details.}
We initialize \oursF\ by copying the \ours's architecture and parameters, except for the learnable tokens and the Gaussian head. We first extract feature maps $\mathbf{F}'_v = \mathcal{E}'(I_v)$ using the user-desired visual encoder $\mathcal{E}'(\cdot)$. We then introduce new learnable feature tokens $\mathbf{Q}'$ and concatenate them with $\mathbf{F}_v$, which are then processed through $\mathcal{T}_\mathcal{F}(\cdot)$ with the same architecture as $\mathcal{T}_\mathcal{G}(\cdot)$:
\begin{equation}
    [\bar{\mathbf{Q}}';\bar{\mathbf{F}}'] = \mathcal{T}_\mathcal{F}([\mathbf{Q}';\mathbf{F}']),
\end{equation}
where $\bar{\mathbf{Q}}'$ denotes the refined learnable tokens and $\bar{\mathbf{F}}'$ denotes the refined features.
In the self-attention layers of $\mathcal{T}_\mathcal{F}$, we reuse query $Q$ and key $K$ features from $\mathcal{T}_\mathcal{G}$ with stop-gradients, while only the value projection layer is newly trained:
\begin{equation}
    V' = [V'_G;V'_I] = \mathcal{P}'_V([\mathbf{Q'};\mathbf{F'}]),
\end{equation}
where $V'_G, V'_I$ denote the value features of the learnable feature tokens and features from $\mathcal{E}'$, respectively, and $\mathcal{P}'_V$ denotes the projection layer of value in $\mathcal{T}_\mathcal{F}$.

Then, the resulting output of self-attention layer in $\mathcal{T}_\mathcal{F}$ is computed as:
\begin{equation}
    \mathsf{Attn}(Q,K,V') = \mathsf{Softmax}(\frac{QK^\intercal}{\sqrt{d_Q}})V'.
\end{equation}

These refined learnable feature tokens $\bar{\mathbf{Q}}'$ are converted to multi-view aggregated features $\mathbf{F}''_i$ for each Gaussian $\mathbf{G}_i$ by passing the feature head $H_\mathcal{F}$, which consists of a single linear layer:
\begin{equation}
    \mathbf{F}''_i = H_\mathcal{F}(\bar{\mathbf{Q}}_i').
\end{equation}

Note that the introduction of new learnable queries $\mathbf{Q}'$ is to enable \oursF\ to take features extracted from any user-desired visual encoder $\mathcal{E}'(\cdot)$ as input, considering that different features~($\mathcal{E}(\cdot), \mathcal{E}'(\cdot)$) will have different feature dimensions. When the two visual encoders are identical~($\mathcal{E}(\cdot) = \mathcal{E}'(\cdot)$), we can additionally copy the learned queries $\mathbf{Q}$ from \ours, only training the value projection layer $\mathcal{P}'_V(\cdot)$ and the feature head $H_\mathcal{F}(\cdot)$.
\section{Experimental settings}
\label{sec:settings}
\subsection{Novel view synthesis}
\paragrapht{Datasets.}
To evaluate novel view synthesis, we use the RealEstate10K~\cite{zhou2018stereo} dataset. We adopt the same train-test split as prior work~\cite{charatan2024pixelsplat, chen2024mvsplat, ye2024no}. RealEstate10K primarily contains indoor real estate videos with camera poses computed using COLMAP~\cite{schonberger2016structure}. For multi-view evaluation, we randomly sample 1,000 image sets from the test split, each consisting of multiple context views and 3 target views that are disjoint from the context views.

\paragrapht{Baselines.}
For 2-view input settings, we compare against SOTA generalizable feed-forward methods on novel view synthesis. We compare PixelSplat~\cite{charatan2024pixelsplat} and MVSplat~\cite{chen2024mvsplat} as \textit{Pose-dependent} categories, which require ground-truth pose information for input. We also compare our model with \textit{Pose-free} categories which use only rgb images as inputs such as CoPoNeRF~\cite{hong2024unifying}, Splatt3R~\cite{smart2024splatt3r}, PF3plat~\cite{hong2024pf3plat}, SPFSplat~\cite{huang2025no}, NoPoSplat~\cite{ye2024no}. We also reimplement the VGGT+NoPo, which replaces NoPoSplat’s MASt3R~\cite{leroy2024grounding} backbone with VGGT~\cite{wang2025vggt} while maintaining NoPoSplat’s pipeline to estimate per-pixel Gaussians. Ours also fall into \textit{Pose-free} settings.

For the multi-view~($V>2$) settings, we compare against state-of-the-art generalizable methods that support arbitrary numbers of input views, such as AnySplat~\cite{jiang2025anysplat}. We evaluate using various numbers of input views: 12, 24, and 36. We also evaluate VGGT+NoPo under the same settings. For AnySplat, we provide input images at $448\times448$ resolutions, which perform better than $256 \times 256$, and render target view images at $256 \times 256$ resolution.

\paragrapht{Evaluation protocol.}
Given unposed images as inputs, our method reconstructs and represents 3D scenes using 3D Gaussians. We rasterize the 3D Gaussians at ground-truth camera poses and compare the rendered images with ground-truth target views. We report standard novel view synthesis metrics such as PSNR, SSIM, and LPIPS.

However, as discussed in NoPoSplat~\cite{ye2024no}, reconstructing 3D scenes from sparse unposed views is inherently scale-ambiguous. Although our method successfully generates 3D Gaussians, they may not fully align with the ground-truth scene scale in the validation dataset. To ensure fair comparison with other baselines, we follow pose-free methods by optimizing the target view camera pose while freezing all other parameters. Specifically, we first reconstruct 3D Gaussians, then freeze them and optimize only the target camera pose such that the rendered image closely matches the ground-truth target view. Finally, we compute metrics using the optimized target view camera poses. Note that this optimization scheme is only needed for evaluation purposes and is not required in real-world scenarios.

For the multi-view settings, we additionally perform short test-time optimization following 3DGS~\cite{kerbl20233d}, denoted as C3G w/ TTO, NoPo+VGGT w/ TTO and AnySplat w/ TTO. We use the predicted Gaussians as initialization and conduct per-scene optimization for a limited number of steps. Specifically, we set the optimization steps to 1,000 steps with densification intervals of 100 steps. We set the learning rate as follows: means at 1.6e-4, scales at 3e-4, rotations at 1e-3, harmonics at 2.5e-3, and densities at 5e-3, using Adam optimizer~\cite{kingma2014adam}.
Our loss function is formulated as $\mathcal{L} = 0.8\mathcal{L}_\text{MSE} + 0.2\mathcal{L}_\text{SSIM}$, where $\mathcal{L}_\text{MSE}$ denotes the MSE loss and $\mathcal{L}_\text{SSIM}$ denotes the SSIM loss between the rendered image and ground-truth images. Other settings follow the original 3DGS. For AnySplat, we follow their default test-time optimization and do not conduct densification to avoid the $\texttt{Out-Of-Memory}$ due to the number of Gaussians.

\subsection{3D scene understanding}

\paragrapht{Datasets.}
To evaluate 3D scene understanding capabilities, we follow previous approaches~\cite{fan2024large, burgess2006spatial} and use the ScanNet~\cite{dai2017scannet} and Replica~\cite{straub2019replica} datasets.

For ScanNet, we use 40 scenes following LSM~\cite{fan2024large}, selected based on valid pose and depth data where Feature-3DGS~\cite{zhou2024feature} performs well. We evaluate 8 categories: wall, floor, ceiling, chair, table, sofa, bed, and others. We select 30 images with a stride of 10, and target views are chosen as the 1st and 4th images within every 8-image interval.

For Replica, we select 80 images with a stride of 3, and the target view is the 2nd image within every 8-image interval. We use camera poses obtained from COLMAP~\cite{schonberger2016structure}. We evaluate 3 scenes where LSeg~\cite{li2022language} performs well: office3, office4, and room1. We select 5-6 categories per scene. Specifically, for office3, we use wall, ceiling, floor, chair, and table; for office4, we use wall, ceiling, floor, chair, tv-screen, and table; for room1, we use wall, ceiling, floor, bed, and blinds.

\paragrapht{Baselines.}
To validate our effectiveness, we compare our model with per-scene optimized feature novel-view synthesis tasks and feed-forward feature novel-view synthesis.

For per-scene optimization methods (Feature-3DGS~\cite{zhou2024feature} and CF$^3$~\cite{lee2025cf3}), we optimize the per-scene 3D Gaussians using all the posed inputs except for target view images. We then render the 3D Gaussians to target view images to calculate the metrics. For the ScanNet dataset, there are no readily available sparse initial point clouds, so we train with randomly initialized point clouds following the previous works~\cite{qin2024langsplat, li2025langsplatv2}. For feature-3DGS~\cite{zhou2024feature}, we optimize for 5,000 steps to avoid overfitting. For CF$^3$~\cite{lee2025cf3}, we first optimize 3DGS with 30,000 steps and additionally optimize the CF$^3$ with 3,000 steps. 

For feature novel-view synthesis (LSM~\cite{fan2024large} and Ours), we first select two input views to generate 3D Gaussians, which are then projected to target views disjoint from the input views. Following the same protocol as novel view synthesis evaluation, these methods require target pose optimization to resolve scale ambiguity. 

For LSeg~\cite{li2022language} and MaskCLIP~\cite{zhou2022extract}, we directly extract features from the target viewpoint images, since these methods cannot render features at novel viewpoints like 3D-based approaches.

\paragrapht{Evaluation protocol.}
Given RGB images as input, we first extract language-aware features using models such as LSeg~\cite{li2022language} or MaskCLIP~\cite{zhou2022extract}. 3D Gaussians with lifted features are generated following each method's procedure, then projected to the target and source views. The target view denotes an unseen viewpoint not present in the input, while the source view denotes a seen viewpoint that every method has observed at least once. We render both RGB images and features at each viewpoint. The rendered features are then processed into segmentation maps by selecting the most relevant CLIP~\cite{radford2021learning} text embeddings corresponding to each label. We report open-vocab segmentation metrics (mIOU and Accuracy), and novel view synthesis metrics (PSNR, SSIM, and LPIPS).

\subsection{Multi-view feature encoding}
Obtaining multi-view invariant features, also termed as 3D aware features, has been a long-standing goal in computer vision and graphics. Probe3D~\cite{el2024probing} defines a set of tasks, including two-view correspondences~\cite{hong2021deep,hong2022cost,cho2021cats,cho2022cats++,hong2024unifying2,hong2022neural} and single-view depth estimation~\cite{piccinelli2024unidepth}, to probe the 3D awareness of the features. FiT3D proposes a two-stage fine-tuning task to build view-invariant features, where they first train more than 1000 per-scene feature-3DGS with the desired features. With the pre-trained 3DGS over multiple scenes, they finetune the original vision encoder to follow the rendered features from the 3D Gaussians. They show that after training, they achieve higher correspondence scores in the task defined in Probe3D. As our \oursF\ can aggregate the input features, we analyze the effectiveness of \oursF\ as a view-invariant feature decoder.

\paragrapht{Datasets.}
To validate the effectiveness of our \oursF\ as a multi-view invariant feature decoder, we use the ScanNet~\cite{dai2017scannet} dataset following Probe3D~\cite{el2024probing}. ScanNet is a large-scale dataset of indoor scenes with RGB images, depth maps, and camera poses. We evaluate on 1,500 image pairs from the test split following SuperGlue~\cite{sarlin2020superglue}.

\paragrapht{Baselines.}
We compare DINOv2~\cite{oquab2023dinov2}, DINOv3~\cite{simeoni2025dinov3}, and VGGT tracking features~\cite{wang2025vggt}, which are known to perform well on zero-shot correspondence estimation.

\paragrapht{Evaluation protocols.}
Given two images, we first extract a feature map for each image. For ours, we first predict 3D Gaussians with features and project the 3D Gaussians to each input-view pose. Then, we estimate correspondences between the two images using nearest neighbors. Following Probe3D~\cite{el2024probing}, we filter the correspondence using Lowe's ratio test~\cite{lowe2004distinctive} to find the strong unique matches to reduce the noisy correspondence. We rank the correspondences using the ratio test and keep the top 1,000 correspondences. 

We evaluate correspondence quality using PCK @ 10px, which measures the accuracy of correspondence within 10 pixels between estimated and ground-truth matches. We divide the image pairs into 4 groups based on viewing angle differences: $0{\sim}15^\circ$~($\theta_0^{15}$), $15{\sim}30^\circ$~($\theta_{15}^{30}$), $30{\sim}60^\circ$~($\theta_{30}^{60}$), and $60{\sim}180^\circ$~($\theta_{60}^{180}$). The results can be seen in Tab.~\textcolor[HTML]{367DBD}{4}.

\subsection{Multi-view feature upsampling}
Feature upsampling is the task of upsampling the extracted features from pre-trained vision encoders to a higher resolution for more fine-grained downstream tasks. Although our \oursF\ also takes the low-resolution extracted features from $\mathcal{E}'(\cdot)$ as input, when combined with the 3D Gaussians and rendered to a specific viewpoint, since Gaussians can be rendered at any camera pose, our model can generate features at arbitrary resolutions by setting the desired resolution and intrinsics in the CUDA rasterizer. In this section, we analyze whether the rendered features from \oursF\ can be used as an effective multi-view feature upsampler.

\paragrapht{Datasets.}
We use the ScanNet~\cite{dai2017scannet} dataset, following the same setup as multi-view feature encoding, to evaluate the probing capability of the upsampled features.

\paragrapht{Baselines.}
We compare DINOv2~\cite{oquab2023dinov2} and DINOv3~\cite{simeoni2025dinov3} features as baselines against their upsampled versions, which match the input image resolution using AnyUp~\cite{wimmer2025anyup}, a generalizable feature upsampling module.

\paragrapht{Evaluation protocols.}
Following the same protocol as multi-view feature encoding, we first extract high-resolution feature maps, then estimate correspondences between two images using nearest neighbors. We evaluate correspondence quality using PCK@10px with image pairs divided by viewing angle differences. We also present qualitative results using PCA visualization to demonstrate how upsampled features capture finer details. The results can be seen in Tab.~\textcolor[HTML]{367DBD}{4}.

\clearpage
\section{Additional ablation and discussion}
\label{sec:ablation_discussion}
\subsection{Comparison of compact 3D reconstruction strategies}
\begin{table}[t]
\centering
\caption{\textbf{Comparison of novel view synthesis with compact Gaussians generation strategies.} }
\label{suptab:coarse_reconstruction}
\resizebox{0.75\linewidth}{!}{
\begin{tabular}{l|cccc}
\toprule
\multirow{2}{*}{Methods}  & \multicolumn{4}{c}{Average} \\
&  PSNR$\uparrow$ & SSIM$\uparrow$ & LPIPS$\downarrow$ & \#G$\downarrow$ \\
\midrule
Sampling & 21.340& 0.665& 0.272& \textbf{2K} \\
Voxelize & 19.957& 0.609& 0.403& 4K \\
Ours & \textbf{22.387}& \textbf{0.713}& \textbf{0.259}& \textbf{2K} \\
\bottomrule
\end{tabular}}
\end{table}
To validate the effectiveness of our compact reconstruction strategy, we conduct ablation studies comparing different approaches to obtain compact representations. For the sampling baseline, we first estimate per-pixel Gaussian centers and their corresponding Gaussians, then downsample by 8$\times$ to predict only 2,048 Gaussians, matching the number of Gaussians with our method. For the voxelization baseline, we follow AnySplat's~\cite{jiang2025anysplat} strategy: we first estimate per-pixel Gaussians, then voxelize them with additional layers. Since AnySplat's original voxel size cannot sufficiently reduce the number of Gaussians, we increase the voxel size to 0.2, which results in approximately 4K Gaussians to best match the 2,048 Gaussians produced by our method.

As shown in Tab.~\ref{suptab:coarse_reconstruction}, our strategy demonstrates superior performance compared to sampling or voxelization methods while using fewer Gaussians than voxelization. Both sampling and voxelization rely on per-pixel estimation, which restricts them to pixel locations and prevents accurate estimation of necessary regions. Additionally, voxelization requires heuristic determination of voxel size and can reduce details in high-frequency regions. In contrast, our approach does not depend on pixel locations as learnable tokens can flexibly estimate Gaussians at appropriate locations, enabling more effective compact Gaussian field reconstruction.

\begin{table}[t]
\centering
\caption{\textbf{Analysis of tradeoff between rendering performance and number of Gaussians.}}
\vspace{-10pt}
\label{sup_tab:tradeoff}
\resizebox{\linewidth}{!}{
\begin{tabular}{l|cccc}
\toprule
\# of Gaussians & PSNR$\uparrow$ & SSIM$\uparrow$ & LPIPS$\downarrow$ & FPS$\uparrow$ \\
\midrule
$ 2048 \times 1$ & 20.625 & 0.623 & 0.321 & \textbf{3083.50} \\
$ 2048 \times 2$ & 20.844 & 0.639 & 0.303 & 3015.03 \\
$ 2048 \times 4$ & 21.080 & 0.644 & 0.297 & 2828.14 \\
$ 2048 \times 8$ & \textbf{21.218} & \textbf{0.653} & \textbf{0.284} & 2476.01 \\
\bottomrule
\end{tabular}}
\end{table}

\subsection{Additional analysis of number of Gaussians}
We first analyze the tradeoff between rendering performance and the number of Gaussians in Tab.~\textcolor[HTML]{367DBD}{6}. Notably, this experiment reveals a slight performance drop when increasing the number of learnable tokens from 2,048 to 4,096. We attribute this to our framework's design of using a fixed number of learnable queries, which encourages each query to learn cross-view correspondences. Increasing the number of tokens appears to weaken this inductive bias toward coherent multi-view aggregation, making attention scores less localized and yielding no additional benefit.
In contrast, our experiments in Tab.~\ref{sup_tab:tradeoff} demonstrate that increasing the number of decoded Gaussians \emph{per} token consistently improves performance, enabling the model to capture fine-grained details from scenes coarsely represented by 2K tokens. We believe this analysis highlights a promising direction for applications that prioritize feed-forward NVS performance. Note that the results in Tab.~\ref{sup_tab:tradeoff} are reported after 75K training steps (out of 450K) to assess feasibility.

\subsection{Additional ablation for feature decoder}
We additionally ablate the components of \oursF. In Tab.~\ref{suptab:addi_abl_comp3df}, we conduct experiments on (1) detaching the copied attention weights from \ours, and (2) propagating feature loss to Gaussian geometry attributes (means, covariances, spherical harmonics, and opacities). When feature loss propagates to the copied attention weights, \ours\ receives ambiguous gradients because the features are not perfectly multi-view consistent. Consequently, the attention mechanism cannot correctly identify correspondences between learnable query tokens and the encoder features $\mathcal{E}(\cdot)$~(e.g., VGGT) for \ours.
To propagate feature loss to Gaussian attributes, we modify the CUDA rasterizer from Feature-3DGS~\cite{zhou2024feature}, which originally propagates feature loss only to Gaussian feature attributes. We extend the CUDA rasterizer to propagate feature loss to all Gaussian attributes. However, this also degrades geometry estimation results because foundation model features are not perfectly multi-view consistent.
We hypothesize that with perfectly multi-view invariant features, feature loss could improve reconstruction quality, especially since photometric loss is also imperfect, as real-world RGB images contain visual artifacts such as appearance variations, lighting changes, and noise.

\begin{table}[t]
\centering
\caption{\textbf{Additional ablation studies for \oursF.} }
\label{suptab:addi_abl_comp3df}
\resizebox{\linewidth}{!}{
\begin{tabular}{cc|ccccc}
\toprule
\multirow{2}{*}{\shortstack{Detach\\attn.}} & \multirow{2}{*}{\shortstack{Detach\\ $\mathcal{L}_\text{feat}$ to $G_i$} } & \multicolumn{5}{c}{Lseg~\cite{li2022language}} \\
& & mIOU$\uparrow$ & Acc.$\uparrow$ & PSNR$\uparrow$ & SSIM$\uparrow$ & LPIPS$\downarrow$ \\
\midrule
\ding{55} & \ding{51} & 0.490& 0.761& 23.078& 0.750&0.293\\
\ding{51} & \ding{55} & 0.490& 0.757& 23.243& 0.754&0.286\\
\textbf{\ding{51}} & \textbf{\ding{51}} & \textbf{0.513}& \textbf{0.783}& \textbf{23.886}& \textbf{0.770}& \textbf{0.285}\\
\bottomrule
\end{tabular}}
\end{table}

\subsection{Discussion of autoencoder in existing methods}
\begin{figure}[t]
    \centering
    \includegraphics[width=\linewidth, height=0.7\linewidth]{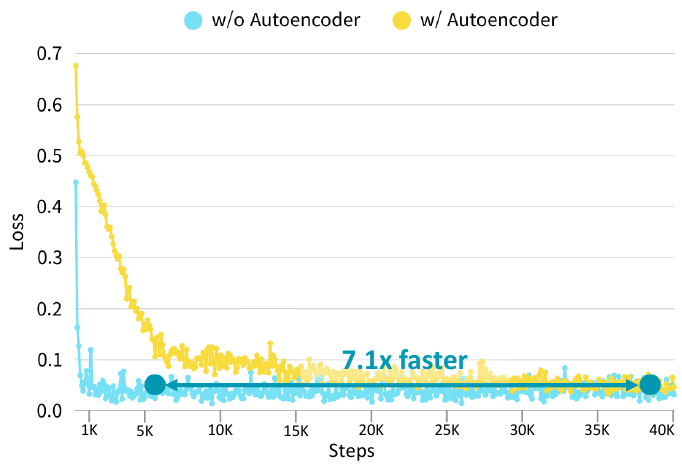}
    \vspace{-15pt}
    \caption{\textbf{Convergence speed improvement by eliminating autoencoder in our framework.}} 
    \label{supfig:convergence}\vspace{-10pt}
\end{figure}

To reduce the memory consumption of Gaussians, prior works~\cite{fan2024large, lee2025cf3} commonly adopt autoencoders to compress Gaussian feature dimensions and restore them via a decoder. However, autoencoder-based compression inevitably introduces information loss.
Moreover, as shown in Fig.~\ref{supfig:convergence}, removing the autoencoder allows our model to converge \textbf{7.1$\times$} faster. With an autoencoder framework, additional training time is required for the encoder–decoder architecture. In contrast, without an autoencoder, we directly leverage the attention mechanism of \ours\ to aggregate and store features. Since the attention maps in \ours\ already identify the necessary features for each learnable token, each feature token naturally attends to the same multi-view regions as its corresponding Gaussian token, effectively reusing learned correspondences for feature aggregation. This emergent property enables highly efficient feature lifting with minimal additional training overhead.

\begin{table}[t]
\centering
\caption{\textbf{Computational Cost.} }
\vspace{-10pt}
\label{sup_tab:computation}
\resizebox{\linewidth}{!}{
\begin{tabular}{c|l|ccc}
\toprule
&\multirow{2}{*}{Methods}  & \multirow{2}{*}{\shortstack{Inference \\ time (s) $\downarrow$}}& \multirow{2}{*}{\shortstack{ GPU Mem. \\ (MiB) $\downarrow$}}& \multirow{2}{*}{FPS $\uparrow$}\\
& & & \\
\midrule
\multirow{3}{*}{\shortstack{2-view\\image\\input}}&NoPoSplat~[\textcolor{cvprblue}{48}] &  0.092& 6,063& 898.28 \\
&NoPo+VGGT & 0.093& 7,495& 1566.28 \\
&Ours & \textbf{0.091}& \textbf{4,150}& \textbf{2742.68} \\
\midrule

\multirow{4}{*}{\shortstack{24-view\\image\\input}} & AnySplat~[\textcolor{cvprblue}{15}] & 1.449& 6,722& 225.50\\
& AnySplat w/ TTO~[\textcolor{cvprblue}{15}] & 35.907& 6,722& 225.50 \\
& Ours & \textbf{0.466}& \textbf{4,372}& \textbf{3083.50} \\
& Ours w/ TTO &  4.273&  \textbf{4,372}& 2511.49 \\
\midrule
\multirow{2}{*}{\shortstack{Feature\\lifting}} & LSM~[\textcolor{cvprblue}{9}] &  0.165& \textbf{5,333}& 625.12\\
&Ours & \textbf{0.093}& 5,896& \textbf{785.35}\\
\bottomrule
\end{tabular}}
\end{table}

\subsection{Discussion of Computational Costs}
We include a detailed analysis of computational costs in Tab.~\ref{sup_tab:computation}. In \textit{novel-view synthesis} settings, compared to prior works~\cite{ye2024no, jiang2025anysplat} that employ a DPT head — which incurs significant computational overhead due to the large feature resolution space — our shallow transformer decoder achieves faster inference with lower peak GPU memory usage. 
For \textit{feature lifting}, our model achieves higher rendering speed and FPS compared to LSM~\cite{fan2024large}. While LSM must compress features to remain tractable due to its large number of Gaussians, our compactness enables feature lifting and rendering without any compression. Although increasing the feature dimension slightly raises peak GPU memory, the overhead is modest and predictable, enabling our method to deliver higher PSNR and mIoU — validating that compact Gaussians provide an efficient and effective substrate for feature lifting.

\subsection{Discussion of FPS}
As shown in Tab.~\textcolor[HTML]{367DBD}{3} and Tab.~\textcolor[HTML]{367DBD}{4}, our method reduces the number of Gaussians by \textbf{65$\times$} (to only \textbf{2K}) compared to per-pixel approaches~\cite{fan2024large,ye2024no}, yet FPS does not scale linearly with this reduction. We further analyze FPS in Tab~\ref{sup_tab:computation}. In the main table, FPS measurements include camera matrix multiplication and tensor operations for pre-processing prior to Gaussian rasterization. We additionally report FPS measured over the rendering step alone, which demonstrates that our compactness becomes more valuable when isolating the rendering stage, yielding substantially higher FPS.

\subsection{Limitations}
Although our view-invariant feature decoder, \oursF, allows lifting arbitrary 2D features into 3D, we have not evaluated all recent foundation models. For instance, we believe that leveraging features from the Segment Anything Model~(SAM)~\cite{kirillov2023segment} could enable robust multi-view consistent segmentation if the features are aggregated within our framework before being decoded by a pre-trained SAM decoder; however, we did not analyze these specific features in this work.

In addition, following prior works, our experimental validation primarily focuses on multi-view open-vocabulary segmentation. To achieve holistic 3D scene understanding, future work could explore integrating our feature fields with recent Multimodal Large Language Models~(MLLMs)~\cite{liu2023visual,liu2024improved} to enable 3D Scene Question Answering. Furthermore, since our framework enables novel view feature rendering without information loss or compression artifacts, it holds significant promise for integration with Vision-Language-Action~(VLA) models or robotics applications. However, such scenarios are frequently dynamic, whereas our current framework is limited to static scene reconstruction. Extending our compact representation to dynamic scenes remains an exciting avenue that would unlock potential across various autonomous fields.

\section{Additional experiment results}
\label{sec:additional_quant}
\subsection{Novel view synthesis via latent decoding}
\begin{figure}[t]
    \centering
    \includegraphics[width=\linewidth]{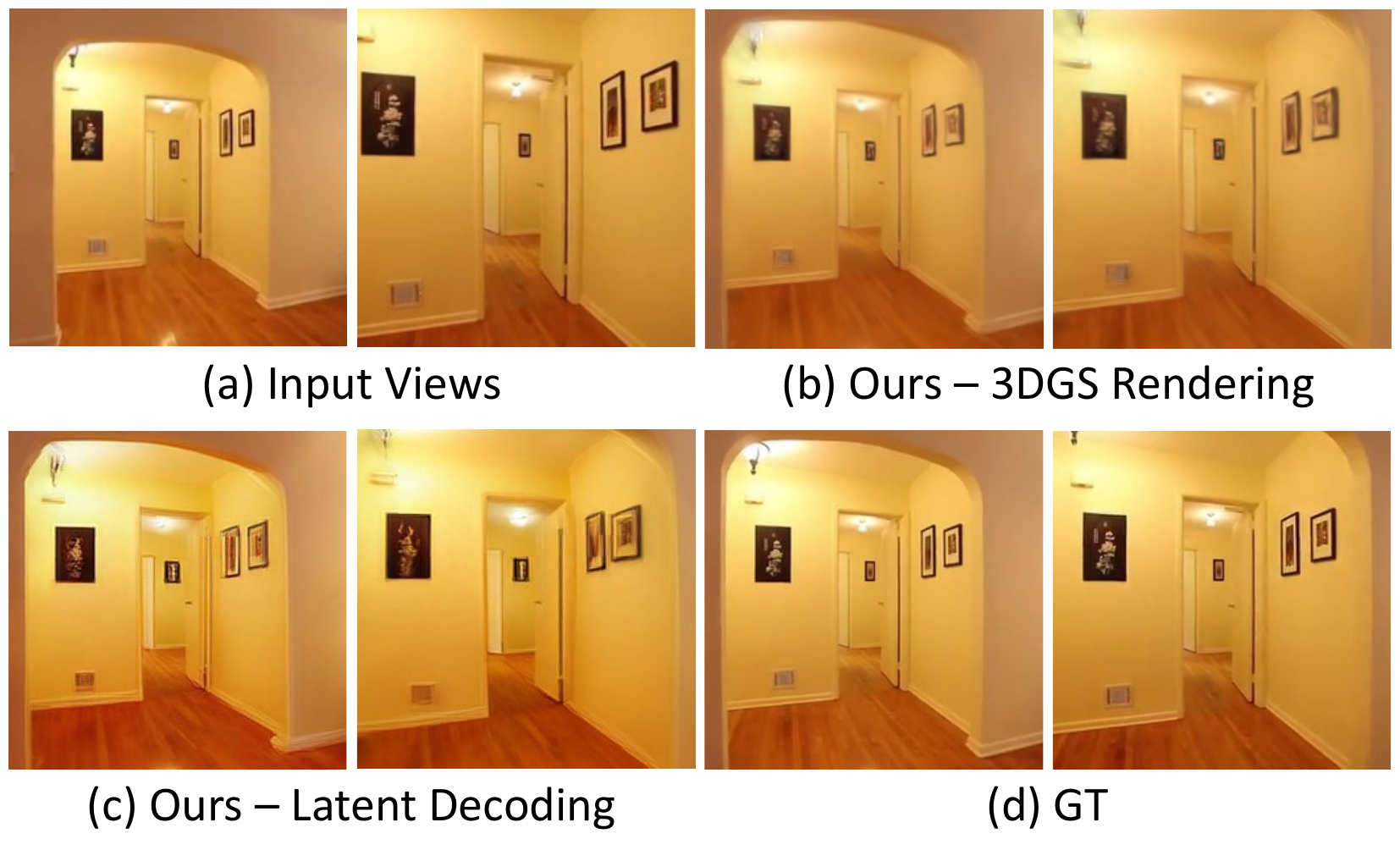}
    \vspace{-15pt}
    \caption{\textbf{Novel view synthesis via latent decoding.} We explore the potential of combining our view-invariant feature decoder, \oursF, with generative models. We lift DINOv2-base features~(which serve as latents for a Representation Autoencoder (RAE)~\cite{zheng2025diffusion}) extracted from the input views~((a)) to 3D Gaussians and render them at novel viewpoints. (b) \textbf{Ours -- 3DGS Rendering}: Standard RGB rendering from the estimated Gaussians provides faithful reconstruction but exhibits some blurriness in high-frequency regions. (c) \textbf{Ours -- Latent Decoding}: By decoding the rendered feature maps through the RAE's diffusion transformer with a 1-step denoising process, we recover significantly sharper textures and edges. Although the generative nature introduces slight variations in fine details compared to the Ground Truth (d), this validates our model's ability to provide consistent 3D-aware latents for diffusion-based pipelines.} 
    \label{supfig:rae}\vspace{-10pt}
\end{figure}

As our multi-view invariant feature decoder \oursF\ can take any desired feature as input, we conduct an additional experiment of leveraging the latents of the representation autoencoder~(RAE)~\cite{zheng2025diffusion} as input. Specifically, we leverage the open-sourced RAE, which learns a diffusion transformer~\cite{peebles2023scalable} and a decoder based on the frozen DINOv2-base~\cite{oquab2023dinov2} as the latent. By lifting the DINOv2-base features and rendering novel viewpoints, we can decode the rendered features using the pre-trained RAE decoder or DiT after adding a small noise to the rendered features. In Fig.~\ref{supfig:rae}, we show the results of novel view synthesis obtained from the estimated 3D Gaussians~(Fig.~\ref{supfig:rae}-(b)) and the novel view synthesis obtained by decoding the rendered features through the DiT~(Fig.~\ref{supfig:rae}-(c)). Specifically, for this experiment, we add a 1-step noise from the original 50-step denoising schedule of RAE-DiT~\cite{zheng2025diffusion}, and denoise with \mbox{1-step}. As shown in Fig.~\ref{supfig:rae}, renderings from 3D Gaussians already provide sufficiently good results but still contain some blurry regions, whereas the latent decoding yields significantly sharper results. However, since the pipeline involves a generative process, even with a single step, we observe slight deformations in fine details. Nevertheless, this experiment demonstrates the potential of combining novel view synthesis with diffusion processes by leveraging latents as input to our \oursF.

\subsection{Novel view synthesis in two-view setting}
In this section, we evaluate novel view synthesis performance on the RealEstate10K dataset~\cite{zhou2018stereo}, following NoPoSplat's~\cite{ye2024no} protocol where two input images are used to estimate 3D Gaussians and render a target view. We compare with both \textit{pose-dependent} models~\cite{charatan2024pixelsplat, chen2024mvsplat} and \textit{pose-free} models~\cite{hong2024pf3plat,hong2024unifying, smart2024splatt3r, ye2024no,huang2025no}. Since we use VGGT~\cite{wang2025vggt} as our visual encoder for \ours, we additionally train VGGT+NoPo, which replaces NoPoSplat's MASt3R~\cite{leroy2024grounding} backbone with VGGT~\cite{wang2025vggt} while maintaining NoPoSplat's pipeline to estimate per-pixel Gaussians. Note that \ours\ directly estimates Gaussians from unposed images, falling into the \textit{pose-free} category. For NoPoSplat, VGGT+NoPo, and ours, we follow NoPoSplat's test-time camera pose optimization, which is only necessary for evaluation. As shown in Tab.~\ref{tab:nvs_baseline}, despite estimating \textbf{65$\times$} fewer Gaussians than per-pixel methods~\cite{hong2024pf3plat,hong2024unifying, smart2024splatt3r, ye2024no,huang2025no}, our approach achieves comparable rendering quality with much faster speeds, validating that our compact Gaussians is sufficient for 3D scene reconstruction.

\begin{table}[t]
\centering
\caption{\textbf{Comparison of novel view synthesis on RealEstate10K~\cite{zhou2018stereo}.} Our method maintains competitive results while using far fewer Gaussians.}
\label{tab:nvs_baseline}
\resizebox{\linewidth}{!}{
\begin{tabular}{c|l|cccccc}
\toprule
\multirow{2}{*}{\shortstack{{Pose-}\\{free}}} & \multirow{2}{*}{Methods} & \multicolumn{6}{c}{Average} \\
& & \#G $\downarrow$ & Memories$\downarrow$ & FPS$\uparrow$ & PSNR$\uparrow$ & SSIM$\uparrow$ & LPIPS$\downarrow$ \\
\midrule
\multirow{2}{*}{\ding{55}} & PixelSplat~\cite{charatan2024pixelsplat} & 131K & 33.6MB &388.03 &  23.848  & 0.806 & 0.185 \\
& MVSplat~\cite{chen2024mvsplat}& 131K & 33.6MB &392.6 & 23.977 & 0.811 & 0.176 \\
\midrule
\multirow{9}{*}{\ding{51}} & CoPoNeRF~\cite{hong2024unifying}  & -& -&0.4 & 18.938 & 0.619 & 0.388\\
& Splatt3R~\cite{smart2024splatt3r} &131K & 33.6MB &393.1& 15.318 & 0.490 & 0.436 \\
& PF3plat~\cite{hong2024pf3plat}&131K & 33.6MB &397.1 & 21.042 & 0.739 & 0.233 \\
& SPFSplat~\cite{huang2025no}&131K & 33.6MB &397.3 & 25.845& 0.852& 0.152\\
& NoPoSplat~\cite{ye2024no} & 131K & 33.6MB &369.8& 25.033& 0.838& 0.160\\
& VGGT+NoPo~\cite{wang2025vggt, ye2024no}& 100K & 25.6MB &419.8 & 23.015& 0.762& 0.187\\
& AnySplat~\cite{jiang2025anysplat} & 320K & 81.9MB & 319.2 & 18.828& 0.656 & 0.358\\
& \textbf{C3G (Ours)} & 2K & 0.1MB & 451.7& 22.387& 0.713& 0.259\\
\bottomrule
\end{tabular}}
\vspace{-10pt}
\end{table}
\section{Additional qualitative results}
\label{sec:additional_qual}
\subsection{Additional results of attention map visualization}
\begin{figure*}[t]
    \centering
    \includegraphics[width=0.9\linewidth]{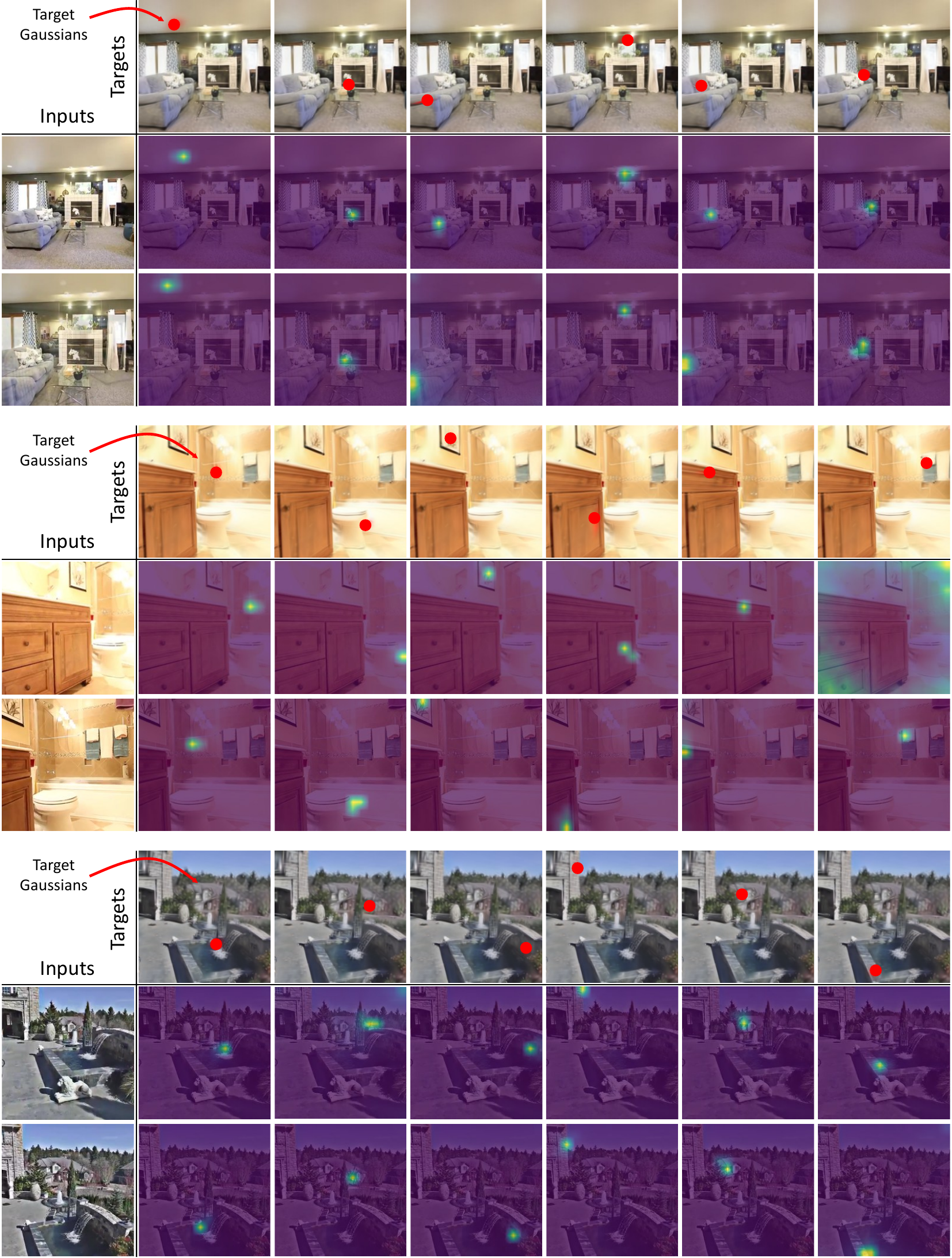}
    \caption{\textbf{Additional visualization of learned attention patterns between a target Gaussian and input images.} Without explicit supervision, each query token (\textcolor{red}{red} dots) learns to attend to spatially coherent regions across multiple views, naturally discovering corresponding regions.} 
    \label{supfig:attention}
\end{figure*}

We additionally visualize the attention maps between learnable tokens $\mathbf{Q}$ and features $\mathbf{F}$ in $\mathcal{T}_\mathcal{G}$, extending Fig.~\textcolor[HTML]{367DBD}{3}-(b). As illustrated in Fig.~\ref{supfig:attention}, \ours\ exhibits sharp, focused attention patterns on spatially coherent regions across multiple views for all Gaussian tokens and input images. These results demonstrate an emergent behavior: to accurately reconstruct novel views with a limited number of $N$ Gaussians, the model learns to position 3D Gaussians at geometrically coherent regions.

\subsection{Additional qualitative results of 3D scene understanding}
\begin{figure*}[t]
    \centering
    \includegraphics[width=\linewidth]{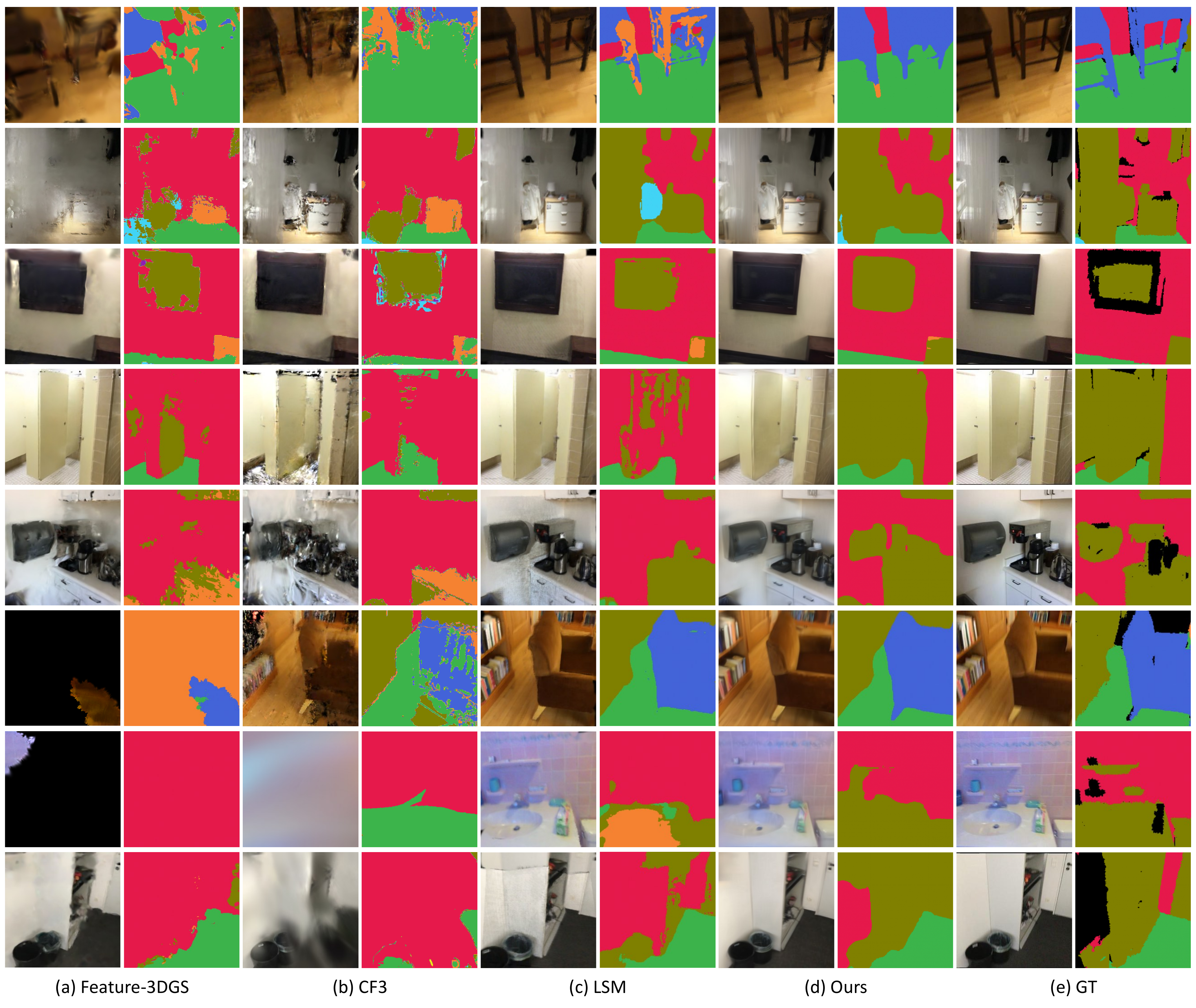}
    \caption{\textbf{Additional qualitative results of 3D scene understanding on ScanNet~\cite{dai2017scannet}.} We conduct qualitative comparison for 3D scene understanding via novel view synthesis and open-vocabulary segmentation. When compared to both per-scene optimization ((a), (b)) and feed-forward ((c), (d)) methods, ours show the most high-fidelity renderings and accurate segmentation maps compared to the ground-truth.} 
    \label{supfig:3d_understanding}
\end{figure*}

We additionally present qualitative results for 3D scene understanding on the ScanNet dataset. In Fig.~\ref{supfig:3d_understanding}, we visualize novel view synthesis results and open-vocabulary segmentation results from feature-lifted 3D Gaussians. Our method generates more geometrically accurate 3D Gaussians and more effectively aggregates multi-view features than competing methods.

\subsection{Qualitative results of multi-view feature encoding}
\begin{figure*}[t]
    \centering
    \includegraphics[width=0.84\linewidth]{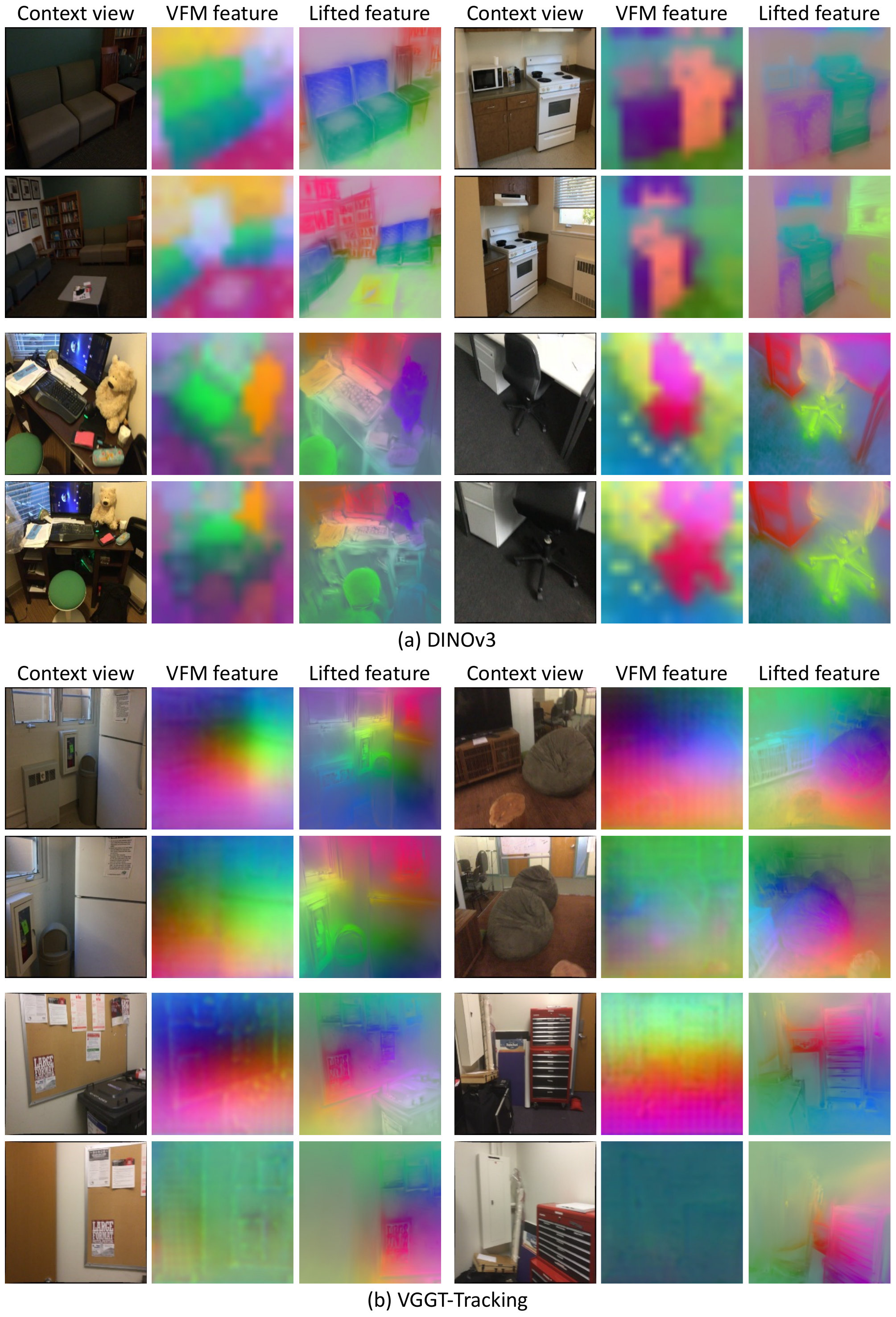}
    \caption{\textbf{Additional PCA visualization of multi-view features on ScanNet~\cite{dai2017scannet}.} We visualize the PCA results of encoded multi-view features. Our method improves multi-view consistency compared to the original visual features.} 
    \label{supfig:mv_pca}
\end{figure*}

In Fig.~\ref{supfig:mv_pca}, we additionally visualize the comparison between pre-lifting features and features aggregated by \oursF\ using PCA. As shown, our model effectively aggregates multi-view features before lifting, producing representations that are both view-invariant and more semantically discriminative.

\subsection{Qualitative results of multi-view feature upsampling}
\begin{figure*}[t]
    \centering
    \includegraphics[width=0.84\linewidth]{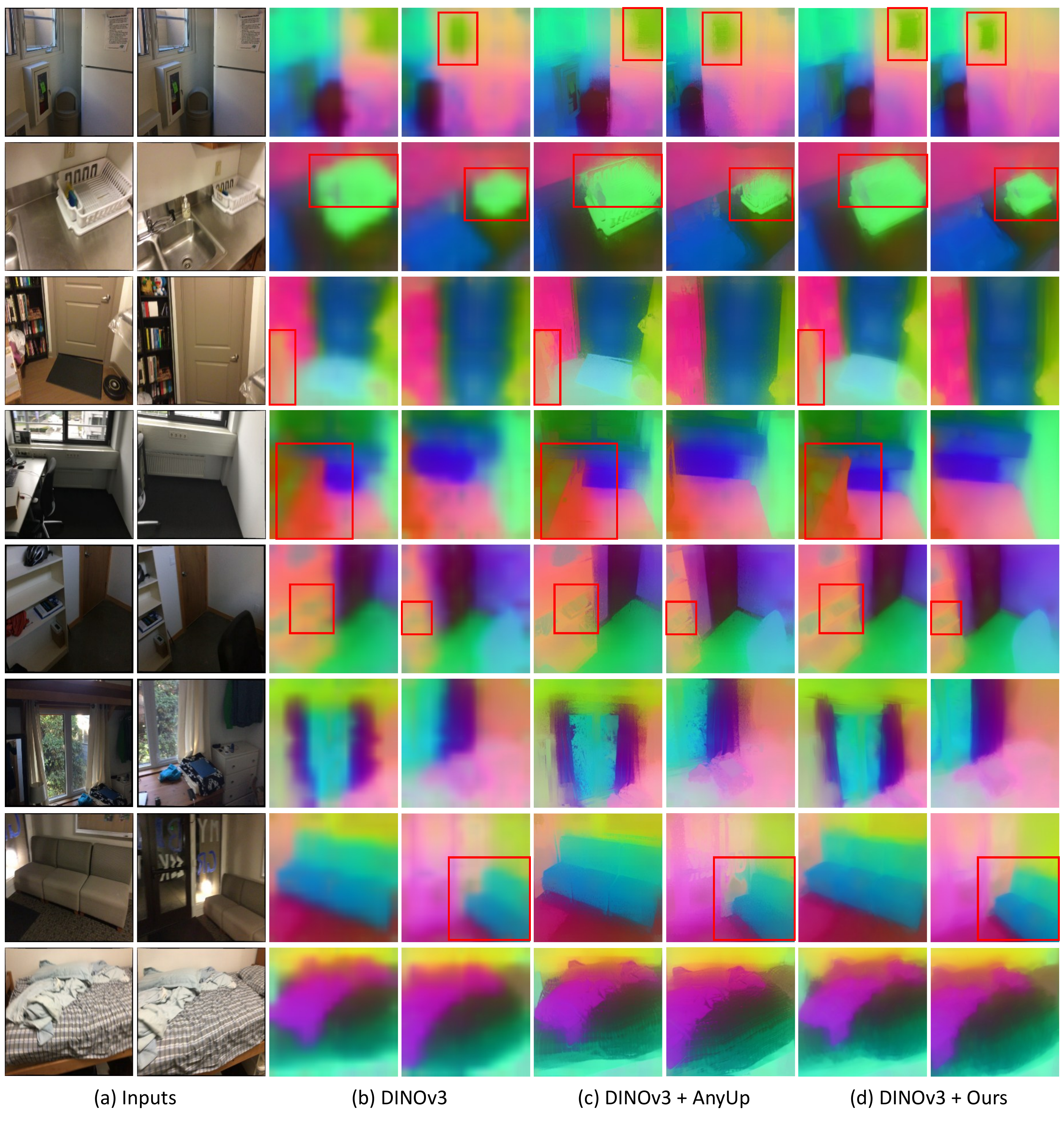}
    \caption{\textbf{Additional PCA visualization of upsampled feature on ScanNet~\cite{dai2017scannet}.} We visualize the PCA results of the upsampled feature. Our model upsamples the features while maintaining the multi-view consistencies compared to other baselines.} 
    \label{supfig:up_pca}
\end{figure*}

In Fig.~\ref{supfig:up_pca}, we present qualitative results of multi-view feature upsampling results. 

\subsection{Qualitative results of novel view synthesis with multi-view inputs}
\begin{figure*}[t]
    \centering
    \includegraphics[width=0.8\linewidth]{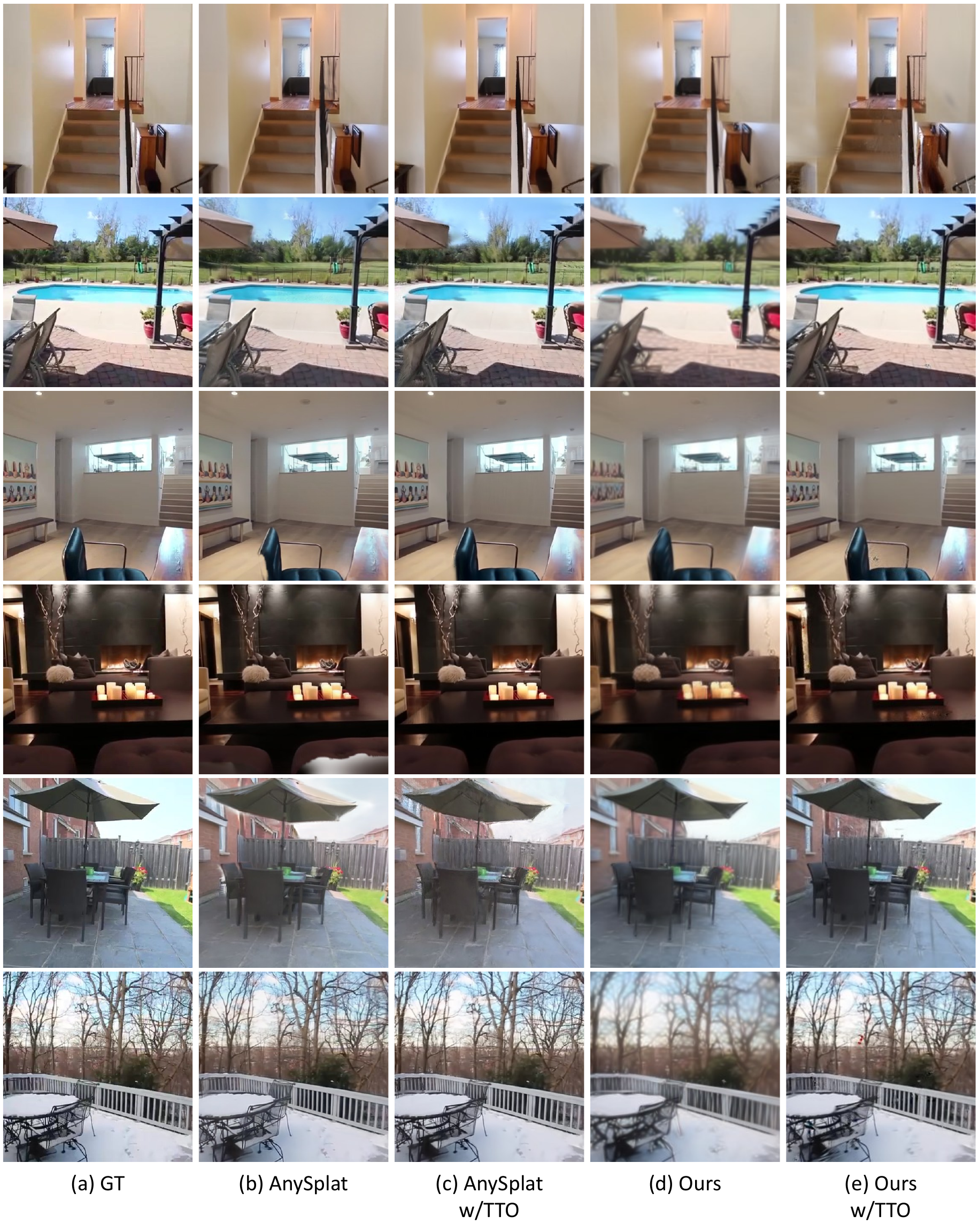}
    \caption{\textbf{Additional qualitative results of novel view synthesis on RealEstate10K~\cite{zhou2018stereo}.} We conduct a qualitative comparison for novel view synthesis with available multi-view images. Our method produces the highest quality rendering results, with or without test-time Gaussian optimization. TTO denotes that test-time optimization is applied to the Gaussians.} 
    \label{supfig:mv_nvs}
\end{figure*}

In Fig.~\ref{supfig:mv_nvs}, we provide qualitative comparisons of novel view synthesis using 24 input images. We compare our method against AnySplat~\cite{jiang2025anysplat}, a state-of-the-art feed-forward approach that supports arbitrary numbers of input views. Despite using significantly fewer Gaussians~(approx. 2K) compared to AnySplat~(approx. 2.6M), our method produces more geometrically consistent renderings with higher quality renderings~((b) vs. (d)). Furthermore, when applying short test-time optimization (denoted as Ours w/ TTO and AnySplat w/ TTO), our method significantly outperforms AnySplat in recovering high-frequency details, demonstrating that our compact representation is sufficient for reconstruction and novel view synthesis while also serving as a robust initialization for per-scene optimization.
\clearpage

\clearpage
{
    \small
    \bibliographystyle{ieeenat_fullname}
    \bibliography{main}
}

\end{document}